\documentclass{article}
\usepackage[numbers]{natbib}
\usepackage[margin=1in,footskip=0.25in]{geometry} 

\usepackage[utf8]{inputenc} 
\usepackage[T1]{fontenc}    
\usepackage{amsfonts}       
\usepackage{nicefrac}       
\usepackage{xcolor}         

\usepackage{amsmath}
\usepackage{amsfonts}
\usepackage{amssymb}
\usepackage{graphicx}
\usepackage{verbatim}
\usepackage{authblk}
\usepackage{graphicx}
\usepackage{doi}
\usepackage{times}
\usepackage{amsthm}
\usepackage{mathrsfs}
\usepackage{authblk}
\usepackage{amsmath,amsfonts,amssymb, graphicx, url} 

\usepackage{caption}
\usepackage{microtype}
\usepackage{booktabs} 
\usepackage{adjustbox}

\usepackage{verbatim}
\usepackage[titletoc]{appendix}
\usepackage{booktabs} 
\usepackage{algorithmic}

\usepackage{algorithm}
\usepackage{amsmath}
\usepackage{amsthm}
\usepackage{amsmath}
\usepackage{esvect}
\usepackage{xcolor}
\usepackage{amsmath,mathtools,amssymb,lipsum}

\DeclareMathOperator{\cov}{cov}

\usepackage{cuted}
\newcommand\numberthis{\addtocounter{equation}{1}\tag{\theequation}}
\usepackage{relsize}

\usepackage{tikz}
\usetikzlibrary{decorations.pathreplacing,calc}

\usepackage[framemethod=TikZ]{mdframed}
\mdfsetup{skipabove=10pt,skipbelow=5pt,roundcorner=4pt}

\definecolor{light-gray}{gray}{0.95}
\usepackage[most]{tcolorbox}
\newtcbox{\mymath}[1][]{%
    nobeforeafter, math upper, tcbox raise base,
    enhanced, colframe=blue!30!black,
    colback=blue!30, boxrule=1pt,
    #1}

\usepackage{enumitem}
\usepackage{wrapfig}

\usepackage{enumitem}
\usepackage{wrapfig}

\usepackage{verbatim}
\usepackage[titletoc]{appendix}
\usepackage{booktabs} 
\usepackage{multirow}
\usepackage{color}
\usepackage{amsthm}
\usepackage{esvect}
\usepackage{xcolor}
\usepackage{amsmath,mathtools,amssymb,lipsum}
\DeclareMathOperator{\mv}{mvm}

\usepackage{cuted}
\usepackage{subcaption}
\usepackage{siunitx}
\usepackage{caption}
\usepackage{makecell}

\newcommand{\change}[1]{{#1}}

\usepackage[T1]{fontenc}
\usepackage[font=small,labelfont=bf,tableposition=top]{caption}

\DeclareCaptionLabelFormat{andtable}{#1~#2  \&  \tablename~\thetable}

\newtheorem*{rep@theorem}{\rep@title}
\newcommand{\newreptheorem}[2]{%
\newenvironment{rep#1}[1]{%
 \def\rep@title{#2 \ref{##1}}%
 \begin{rep@theorem}}%
 {\end{rep@theorem}}}
\newtheorem{theorem}{Theorem}
\newtheorem{proposition}{Proposition}

\newtheorem{claim}{Claim}
\newreptheorem{theorem}{Theorem}
\newreptheorem{claim}{Claim}
\newreptheorem{proposition}{Proposition}
\newcommand{\norm}[1]{\left\lVert#1\right\rVert}
\newcommand{\R}{\mathbb{R}}

\newcommand{\bv}[1]{\mathbf{#1}}

\newcommand{\cardinalitynorm}[1]{|#1|}
\newcommand{\definedas}{\coloneqq}

\newcommand{\gridset}{\mathcal{G}}
\newcommand{\x}{\mathbf{x}}

\renewcommand{\O}{\mathcal{O}}
\DeclareMathOperator{\vectorize}{vec}
\DeclareMathOperator{\matricize}{mat}

\usepackage{bm}

\usepackage{multirow}
\usepackage{array}
\usepackage{url}

\usepackage{tikz}
\usetikzlibrary{decorations.pathreplacing,calc}

\usepackage[T1]{fontenc}
\usepackage[font=small,labelfont=bf,tableposition=top]{caption}

\DeclareCaptionLabelFormat{andtable}{#1~#2  \&  \tablename~\thetable}

\usepackage[sc,osf]{mathpazo}

  \usepackage{nth}
  \usepackage{intcalc}

\usepackage[framemethod=TikZ]{mdframed}
\mdfsetup{skipabove=10pt,skipbelow=5pt,roundcorner=4pt}

\usepackage{enumitem}
\usepackage{wrapfig}

\usepackage{cuted}
\hypersetup{
    colorlinks=true,
    citecolor=black,   
    urlcolor=black,
    linkcolor=black
}

\title {Kernel Interpolation with Sparse Grids}

\author{Mohit Yadav, Daniel Sheldon, Cameron Musco \\
\normalsize{University of Massachusetts Amherst} \\ 
\normalsize{\texttt{\{ymohit, sheldon, cmusco\}@cs.umass.edu}}}
\date{}

\begin{document}
\maketitle
\begin{abstract}
  Structured kernel interpolation (SKI) accelerates Gaussian process (GP) inference by interpolating the kernel covariance function using a dense grid of inducing points, whose corresponding kernel matrix is highly structured and thus amenable to fast linear algebra.
  Unfortunately, SKI scales poorly in the dimension of the input points, since the dense grid size grows exponentially with the dimension. To mitigate this issue, we propose the use of \emph{sparse grids} within the SKI framework. These grids enable accurate interpolation, but with a number of points growing more slowly with dimension. 
  We contribute a novel nearly linear time matrix-vector multiplication algorithm for the sparse grid kernel matrix. 
  Next, we describe how sparse grids can be combined with an efficient interpolation scheme based on simplices. 
  With these changes, we demonstrate that SKI can be scaled to higher dimensions while maintaining accuracy.
 \end{abstract}

\section{Introduction}
\label{sec:introduction}
Gaussian processes (GPs) are popular prior distributions over continuous functions for use in  Bayesian inference \citep{neal}.
Due to their simple mathematical structure, closed form expressions can be given for posterior inference \citep{rasmussen2004gaussian}. 
Unfortunately, a  well-established limitation of GPs is that they are difficult to scale to large datasets. 
\change{In particular, for both exact posterior inference and 
the exact log-likelihood computation for hyperparameter learning, one must 
invert a dense kernel covariance matrix $K \in \R^{n \times n}$,  where $n$ is the number of training points. Naively, this operation requires $\O(n^3)$  time and $\O(n^2)$  memory. }

\medskip

\noindent\textbf{Structured Kernel Interpolation.} Many techniques have been 
proposed to mitigate this scalability issue 
\cite{Snelson2005SparseGP, rahimi2008random,titsias2009variational,GPyTorchBM}. 
Recently, structured kernel interpolation (SKI) has emerged as a promising approach \citep{GPyTorchBM}. 
In SKI, the kernel matrix is approximated via interpolation onto a dense rectilinear grid of $m$ inducing points. In particular, $K$ is approximated as $W K_G W^T$, 
where $K_G \in \R^{m \times m}$ is the kernel matrix on the inducing points and $W \in \R^{n \times m}$ is an interpolation weight matrix mapping training points to nearby grid points. 
Typically, $W$ is sparse, and $K_G$ is highly structured — e.g., for shift invariant kernels, $K_G$ is multi-level Toeplitz. 
Thus, $W$, $K_G$, and in turn the approximate kernel matrix $W K_G W^T$ admit fast matrix-vector multiplication. 
This allows fast approximate inference and log-likelihood computation via 
the use of iterative methods, e.g., the conjugate gradient algorithm.

\medskip

\noindent\textbf{SKI's Curse of Dimensionality.} 
Unfortunately, SKI does not scale well to high-dimensional input data: the number of points in the dense grid, and hence the size of $K_G$,
 grows exponentially in the dimension $d$. Moreover, SKI typically employs local cubic interpolation, which leads to 
 an interpolation weight matrix $W$ with row sparsity that also scales exponentially in $d$. This  \textit{curse of dimensionality} is a well-known issue 
 with the use of dense grid interpolation. It has been studied extensively, e.g., in the context of high-dimensional interpolation
  and numerical integration  \citep{phillips1996theory, Dutt1996FastAF}. 

\begin{figure}[h!]
\begin{minipage}{0.45\columnwidth}
\centering
\includegraphics[width=\columnwidth,height=64mm]{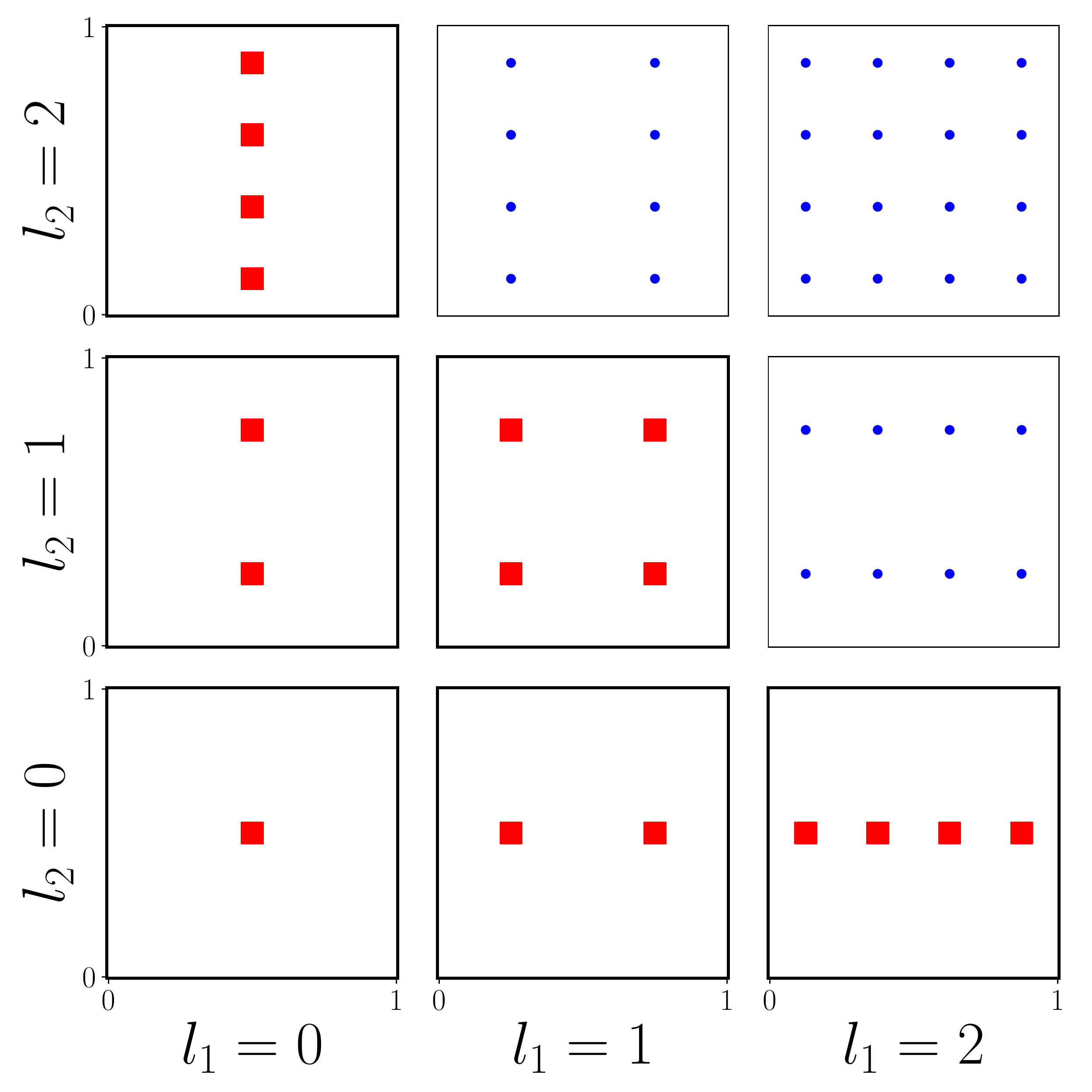} 
\par
(a)
\end{minipage}
\begin{minipage}{0.52\columnwidth}
\centering
\includegraphics[width=0.4\columnwidth,height=26mm]{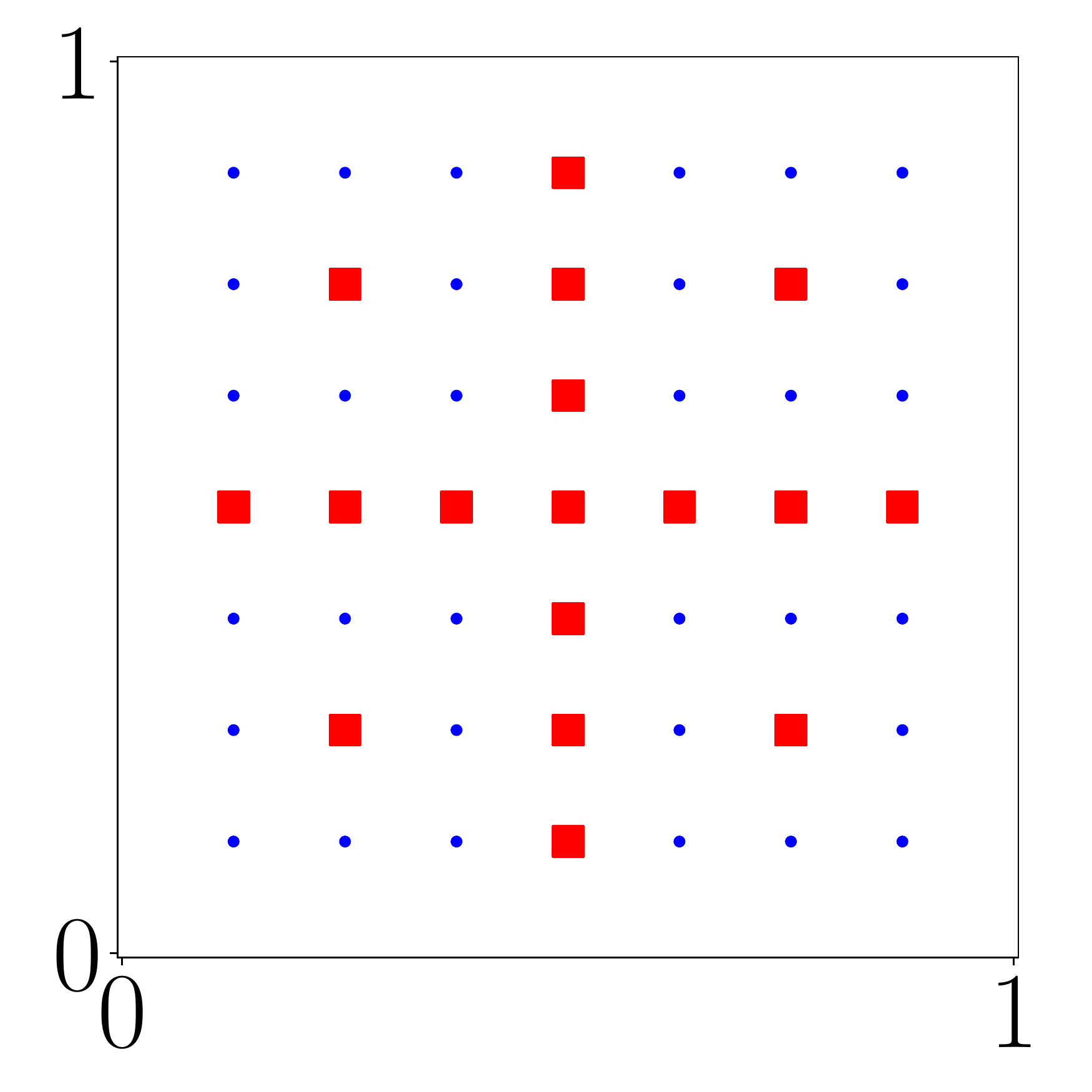} 
\par
(b)
\\[1mm]
\centering
\includegraphics[width=0.7\columnwidth, height=32mm]{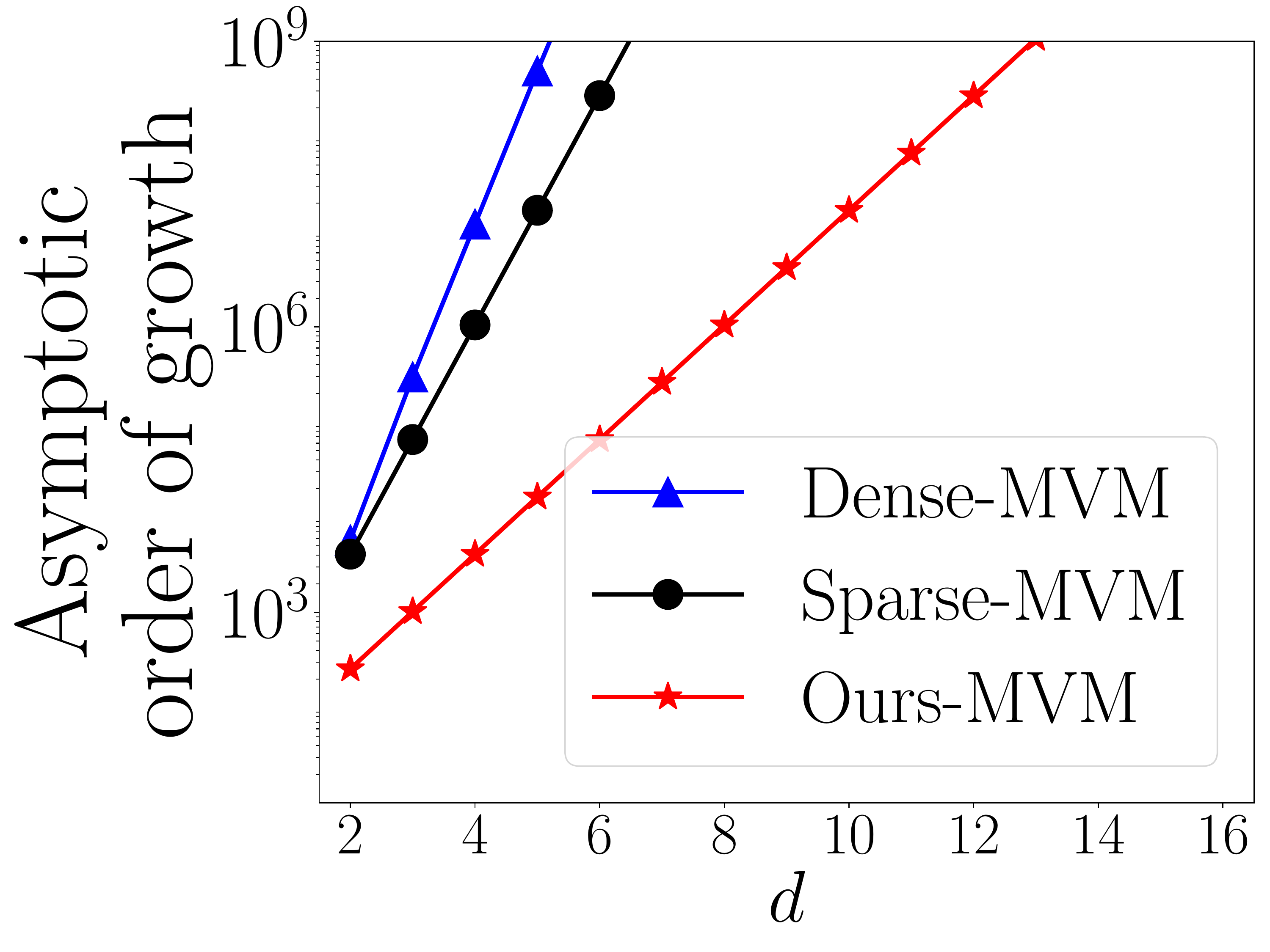}
\par
(c)
\end{minipage}
\caption{Illustration of sparse grid construction for $d=2$ and maximum resolution $\ell=2$.
(a) A dense grid with resolution $(l_1, l_2)$ has $2^{l_1}$ and $2^{l_2}$ points in each dimension; 
\change{the grids with red square points  have total resolution $l_1 + l_2 \leq 2$}.
(b) The sparse grid $\gridset_{2, 2}$ is the union of rectilinear grids with total resolution at most 2. The 17 points with red squares belong to the sparse grid. The dense grid includes the additional points with blue dots, for a total of 49 points.
(c) \change{The asymptotic order of growth in performing a single kernel MVM operation for both grids with $2^8$ unique points 
 in each dimension, ignoring constants.} For sparse grids, \emph{Ours-MVM} (i.e., the proposed MVM algorithm)  improves significantly over the naive implementation (\emph{Sparse-MVM}). 
}
\label{fig:sparse_grid_construnction_and_mvm_times}
\end{figure}

In the computational mathematics 
community, an important technique for interpolating functions in high dimensions is \emph{sparse grids} \citep{bungartz_griebel_2004}. Roughly a sparse grid is a union of rectilinear grids with different resolutions in each dimension. In particular, it is a union of all $2^{\ell_1} \times 2^{\ell_2} \times \ldots \times 2^{\ell_d}$ sized grids, where $\sum_{i=1}^d \ell_i \le \ell$, for some maximum total resolution $\ell$. This upper bound on the total resolution limits the number of points in each grid — while a grid can be dense in a few dimensions, no grid can be dense in all dimensions. See Figure \ref{fig:sparse_grid_construnction_and_mvm_times} for an illustration.
Sparse grids have 
interpolation accuracy comparable to dense grids under certain smoothness assumptions on the interpolated function \citep{smolyak1963quadrature}, while using significantly fewer points.
Concretely, for any function $f: \R^{d} \to \R$ with bounded mixed partial derivatives,  a {sparse grid} containing $\O(2^\ell {\ell}^{d-1})$ points can interpolate as accurately as a dense grid with $\O(2^{\ell d})$  points, where $\ell$ is the maximum grid resolution 
\citep{sickel2011spline}.

\medskip

\noindent\textbf{Combining Sparse Grids with SKI.} 
Our main contribution is to demonstrate that sparse grids can be used within 
the SKI framework to significantly improve scaling with dimension. 
Doing so requires several algorithmic developments. 
When the inducing point grid is sparse, the kernel matrix on the grid,
 $K_G$, no longer has simple structure. E.g., it is not Toeplitz when 
 the kernel is shift invariant. Thus, naive matrix-vector multiplications with 
 $K_G$ require time that scales quadratically, rather than near-linearly in the grid size. This would significantly limit the scope of performance improvement from using a sparse grid. 
To handle this issue, we develop a 
near-linear\footnote{\change{`Near-linear' here means running in time $\O(m \log m)$ for a sparse grid with $m$ points.}} time matrix-vector multiplication (MVM) algorithm for any sparse grid kernel matrix. 
Our algorithm is recursive, and critically leverages the fact that sparse grids 
can be constructed from smaller dense grids and that they have tensor product structure across dimensions. 
For an illustration of our algorithm's complexity versus that of naive quadratic time MVMs, see Figure \ref{fig:sparse_grid_construnction_and_mvm_times} (c).

A second key challenge is that, while sparse grids allow for a grid size that 
grows as a much more mild exponential function of the dimension $d$, 
the bottleneck for applying SKI on large datasets can come in computing MVMs 
with the interpolation weight matrix $W \in \R^{n \times m}$. 
For classic high-dimensional interpolation schemes, like cubic interpolation, 
each row of $W$ has $\O(2^{d})$ non-zero entries, i.e., the kernel covariance for 
each training point is approximated by a weighted sum of the covariance 
at $\O(2^{d})$ grid points. When the number of training points $n$ is large, 
storing $W$ in memory, and multiplying by it, can become prohibitively expensive. 
To handle this issue, we take an approach similar to that of  
\citet{SimplicesGP} and  employ simplicial basis functions for interpolation, whose support grows linearly with $d$.
Combined with our fast MVM algorithm for sparse grid kernel matrices, this interpolation scheme 
lets us scale SKI to higher dimensions. 

In summary, we propose the use of \textit{sparse grids} to improve the scalability of
kernel interpolation for GP inference relative to the number of dimensions. 
To this end, we develop an efficient nearly linear time matrix-vector multiplication algorithm for the sparse grid kernel matrix.
Furthermore, we also propose the use of simplicial interpolation to improve scalability of SKI for both dense and sparse grids.
We show empirically that these ideas allow SKI to scale to at least 10 dimensions and perform competitively with state-of-the art GP regression methods.
We provide an efficient GPU implementation of the proposed 
algorithm  compatible with GPyTorch \citep{GPyTorchBM}, 
which is available at \url{https://github.com/ymohit/skisg} and licensed under the MIT license.

\section{Background}
\label{sec:background}

\noindent\textbf{Notation.} We let 
$\mathbb{N}$ and $\mathbb{N}_0$ denote the natural numbers and  $\mathbb{N} \cup \{0\}$ respectively. 
Matrices are represented by capital letters, and vectors by bold letters. $I$ denotes the identity matrix, with dimensions apparent from context.
For a matrix $M$, $\mv(M)$ denotes the number of operations required to multiply $M$ by any admissible vector. 
In GP regression, the training data are modeled as noisy measurements of a random function $f$ drawn from a GP prior, denoted $f \sim \mathcal{N}(0, k(\cdot,\cdot))$, where $k: \R^d \times \R^d \rightarrow \R$ is a covariance kernel. For an input $\x_i$, the observed value is modeled as $\bv{y}_i = f(\bv{x}_i) + \epsilon_i$, with $\epsilon_i \sim  \mathcal{N}(0, \sigma^2)$.

Observed training pairs $(\x_i, y_i)$ are collected as $X = [ \bv{x_1}, \ldots, \bv{x_n} ] \in \R^{n \times d}$ and $\bv{y} = [y_1, \ldots, y_n] \in \R^{n}$. The kernel matrix (on training data) is $K_{X} = [k(\mathbf{x}_i,\mathbf{x}_j)]_{i,j=1}^{n} \in \mathbb{R}^{n \times n}$.
The GP inference tasks are to compute the posterior distribution of $f$ given 
$(X, \bv y)$, which itself is a Gaussian process, and to compute the
 marginal log-likelihood $\log p(\bv y)$. Naive approaches rely on the Cholesky decomposition of the matrix $\overline{K}_X = {K}_{X}+\sigma^2 I$, which takes $\Theta({n^3})$ time; see~\citet{rasmussen2004gaussian} for more details.

To avoid the $\Theta({n^3})$ running time of naive inference, many modern methods use iterative algorithms such as the conjugate gradient (CG) algorithm to perform GP inference in a way that accesses the kernel matrix only through matrix vector multiplication (MVM), i.e., the mapping $\bf{v} \mapsto$~$ \overline{K}_X \bf{v}$~\citep{GPyTorchBM}.
These methods support highly accurate approximate solutions to GP posterior inference task as well as hyper-parameter optimization.
The complexity of posterior inference and one step of hyper-parameter optimization 
is $\Theta(pn^2)$ for $p$ CG iterations, as $\mv(\overline{K}_X)=n^2$.
In practice, $p \ll n$ suffices~\citep{kissgp, GPyTorchBM}. 

\subsection{SKI: Structured Kernel Interpolation}
\label{ssec:ski}

SKI further accelerates iterative GP inference by approximating the kernel matrix in a way that makes matrix-vector multiplications faster~\cite{kissgp}. 
Given a set of inducing points $U \subset \R^{d}$, SKI approximates the kernel function as $\tilde k(\bv{x},\bv{x'}) \triangleq \bv{w}_{\bv{x}}^T K_{U} \bv{w}_{\bv{x}'}$, where $K_{U} \in \R^{\cardinalitynorm{U} \times \cardinalitynorm{U}}$ is the kernel matrix for the set of inducing points $U$, and
the vector $\bv{w}_\bv{x} \in \R^{\cardinalitynorm{U}}$ contains 
interpolation weights to interpolate from $U$ to any $\bv x$.
The SKI approximate kernel matrix is $\tilde K_{X} = W K_{U} W^T$, where $W \in \R^{n \times \cardinalitynorm{U}}$ is the matrix with $i^{th}$ row equal to $\bv{w}_{\bv{x}_i}$.

To accelerate matrix-vector multiplications with the approximate kernel matrix, SKI places inducing points on a regular grid and uses grid-based interpolation. This leads to a sparse interpolation weight matrix $W$ -- e.g., with cubic interpolation there are $O(4^d)$ entries per row, so that $\mv(W) = O(n 4^d)$ -- and to a kernel matrix $K_U$ that is multi-level Toeplitz (if $k$ is stationary)~\cite{kissgp}, so that $\mv(K_U) = O(|U| \log |U|)$. Overall, $\mv(\tilde K_X) = O(n 4^d + |U|\log |U|)$, which is much faster than $n^2$ for small $d$.
However, SKI becomes infeasible in higher dimensions due to the $4^d$ entries per row of $W$ and curse of dimensionality for the number of points in the grid: specifically, $|U| = m^d$ for a grid with $m$ points in each dimension.

\subsection{Sparse Grids}
\label{ssec:sparse_grid_interpoaltion}

\paragraph{Rectilinear grids.}
We first give a formal construction of rectilinear grids, which will later be the foundation for sparse grids~\citep{garcke2012sparse}. 
For a resolution index $l \in \mathbb{N}_0$, define the $1$-d grid $\Omega_{l}$ as the centers of $2^l$ equal partitions of the interval $[0, 1]$, which gives $\Omega_{l} \definedas \{ i/2^{l+1} \, \vert\, 1 \leq i \leq 2^{l+1} \text{ and $i$ is \textbf{ odd}}\}$.
The fact that the position index $i$ must be odd implies that grids for any two different resolutions are \emph{disjoint}.\footnote{Suppose $i/2^{l+1} = j/2^{k+1}$ are both grid points and $k > l$. Then $i=j2^{k-l}$ is even, a contradiction.}
Moreover, resolution-position index pairs $(l, i)$ uniquely specify grid points in the union $\bigcup_{l \in \mathbb{N}_0} \Omega_{l}$ of $1$-d rectilinear grids. 

To extend rectilinear grids to $d$ dimensions, let $\bv{l} \in \mathbb{N}_0^{d}$ 
denote a resolution vector. The corresponding rectilinear grid is given by $\Omega_{\bv{l}} \definedas \otimes_{j=1}^{d} \Omega_{\bv{l}_j}$, where $\otimes$ denotes the Cartesian product.
A grid point in $\Omega_{\bv{l}}$ is indexed by the pair $(\bv{l}, \bv{i})$ of a resolution vector $\bv l$ and \emph{position vector} $\bv i$, where $(l_j, i_j)$ gives the position in the $1$-d grid $\Omega_j$ for dimension $j$.
This construction of rectilinear grids yields three essential properties that will facilitate formalizing sparse grids: (1) a grid $\Omega_{\bv l}$ is uniquely determined by its resolution vector $\bv l$, (2) grids $\Omega_{\bv l}$ and $\Omega_{\bv l'}$ with different resolution vectors are disjoint, (3) the size $\cardinalitynorm{\Omega_{\bv{l}}} = 2^{\norm{\bv{l}}_1}$ of a grid is determined by the $L_1$ norm of its resolution vector.

\medskip

\noindent\textbf{Construction of Sparse Grids.} Sparse grids use rectilinear grids as their fundamental building block \cite{smolyak1963quadrature} and exploit the fact that resolution vectors uniquely identify different rectilinear grids.
Larger grids are formed as the union of rectilinear grids with different resolution vectors.
\emph{Sparse} grids use all rectilinear grids with resolution vector having $L_1$ norm below a specified threshold.
Formally, for a resolution index $\ell \in \mathbb{N}_0$, the sparse grid $\gridset_{\ell, d}$ in $d$ dimensions is $\gridset_{\ell, d} \definedas \bigcup_{\bv l: \norm{\bv l}_1 \leq \ell} \Omega_{\bv l}$.
Figure \ref{fig:sparse_grid_construnction_and_mvm_times} illustrates the construction of the sparse grid  $\gridset_{2, 2}$ 
from smaller 2-d rectilinear grids with maximum resolution $2$, i.e., $\{ \Omega_{\bv{l}} \mid  \norm{\bv{l}}_1 \leq 2 \}$.  
The figure also illustrates another important fact: the sparse grid $\gridset_{\ell, d}$ has a total of $\Theta(2^\ell)$ distinct and equally spaced coordinates in each dimension, but many fewer total points than a dense $d$-fold Cartesian product of such 1-d grids, which would have $\Theta(2^{\ell d})$ points. 

Sparse grids have a number of formal properties that are useful in algorithms and applications~\cite{smolyak1963quadrature,garcke2012sparse}.
Proposition~\ref{proposition:sparse_grid_properties} below summarizes the most relevant ones for our work.
For completeness, a proof appears in appendix~\ref{app:background}.
For more details, see \citet{valentin2019b}. 

\begin{mdframed}[backgroundcolor=light-gray] 
\begin{proposition}[\textbf{Properties of Sparse Grid}]\label{proposition:sparse_grid_properties}
  Let $\gridset_{\ell, d} \subset [0, 1]^d$ be a sparse grid with any resolution $\ell \in \mathbb{N}_{0}$ and dimension $d\in \mathbb{N}$. Then the following properties hold:
\begin{align*}
    &\textsc{(P1)} \,\,\,  \cardinalitynorm{\gridset_{\ell}^{d}} = \O(2^\ell \ell^{d-1}), \\
    &\textsc{(P2)} \,\,\, \forall \ell^{'} \in \mathbb{N},  0 \leq \ell^{'} \leq \ell \,\,\, \implies  \gridset_{\ell^{'},d} \subseteq \gridset_{\ell, d}, \\
    &\textsc{(P3)} 
     \,\,\, \gridset_{\ell, d} = \bigcup {}_{i=0}^{\ell} \big(\Omega_i \otimes \gridset_{\ell-i, d-1}\big) \text{ and } \,\,\, \gridset_{\ell, 1} = \bigcup {}_{i=0}^{\ell} \Omega_{i}.
\end{align*}
\end{proposition}
\end{mdframed}

Property \textsc{P1} shows that the size of sparse grid with $\Theta(2^\ell)$ points in each dimension grows more slowly than a dense grid with the same number of points in each dimension, since $\O(2^\ell \ell^{d-1})  \ll  \O( 2^{\ell d})$. 
Properties \textsc{P2} and \textsc{P3} are consequences of the structure of the set $\left\{ \norm{\bv{l}}_1 \leq \ell \right\}$ and the sparse grid construction.
Property \textsc{P2} says that a sparse grid with smaller resolution is contained in one with higher resolution.
Property \textsc{P3} is a crucial property, and says that a $d$-dimensional sparse grid can be constructed 
via Cartesian products of 1-dimensional dense grids with sparse grids in $d-1$ dimensions. 

\section{Structured Kernel Interpolation on Sparse Grids}
\label{sec:proposed_method}

To scale kernel interpolation to higher dimensions,
 we propose to select inducing points $U = \gridset_{\ell, d}$ on a sparse grid 
 and approximate the kernel matrix as $W K_{\gridset_{\ell,d}} W^T$ for a suitable 
 interpolation matrix $W$ adapted to sparse grids. This will require 
 fast matrix-vector multiplications with the sparse grid kernel matrix 
 $K_{\gridset_{\ell, d}}$ and the interpolation matrix $W$. We show how to 
 accomplish these two tasks in Sections \ref{ssec:fast_mvm_algorithm} and  
 \ref{ssec:interpolation_and_gaussian_processes} 
 \change{ for the important case of stationary product kernels  \cite{gardner2018product}.
}

\subsection{Fast Multiplication with the Sparse Grid Kernel Matrix}
\label{ssec:fast_mvm_algorithm}

Algorithm~\ref{alg:sgkma_algorithm} is an algorithm to compute $K_{\gridset_{\ell,d}} \bv{v}$ for any vector $\bv{v}$. 
The algorithm uses the following definitions. For any finite set $U$, let $K_U = \big[k(\x, \x')\big]_{\x, \x' \in U}$. 
The rows and columns of $K_U$ are ``$U$-indexed'', meaning the entries correspond to elements of $U$ under some arbitrary fixed ordering.
 For $U \subseteq V$, we introduce a selection matrix $\mathcal S_{U, V}$ to map between $U$-indexed and $V$-indexed vectors. 
 It has entries $(\mathcal S_{U,V})_{ij}$ equal to one if the $i$th element of $U$ is equal to the $j$th element of $V$, and zero otherwise. 
Also, $\mathcal S_{V,U} \definedas \mathcal S_{U, V}^T$. For a $V$-indexed vector $\mathbf z_V$, the multiplication $\mathcal S_{U,V} \mathbf z_V$ produces a $U$-indexed vector by selecting entries corresponding to elements in $U$, and for a $U$-indexed vector $\mathbf z_U$, the multiplication $\mathcal S_{V,U} \mathbf z_U$ produces a $V$-indexed vector by inserting zeros for elements not in $U$.

\renewcommand{\algorithmicrequire}{\textbf{Input:}}
\renewcommand{\algorithmicensure}{\textbf{Output:}}
\begin{figure}[ht]
  \begin{mdframed}[backgroundcolor=light-gray]
    \vspace{-10pt}
\begin{algorithm}[H]
\caption{Sparse Grid Kernel-MVM Algorithm}\label{alg:sgkma_algorithm}
\begin{algorithmic}[1]
\REQUIRE $\bv{v} \in \R^{\cardinalitynorm{\gridset_{\ell,d}}}$ and $K_{\gridset_{\ell,d}} \in \R^{\cardinalitynorm{\gridset_{\ell,d}} \times \cardinalitynorm{\gridset_{\ell,d}}}$
\ENSURE \change{$\bv{u} =  \textbf{mvm}  \left(K_{\gridset_{\ell, d}}, \bv{v}\right)$, where, 
$\textbf{mvm}  \left(K, \bv{v}\right)$ denotes $K \bv{v}$ obtained using this algorithm. \\[2pt]}
\STATE Let $V_i$ be the result of reshaping $\mathcal S_{\Omega_i \otimes \gridset_{\ell-i, d-1}, \gridset_{\ell, d}}\, \bv v$ into a 
$|\Omega_i| \times |\gridset_{\ell-i, d-1}|$ matrix; this contains entries of $\bv v$ corresponding to the $i$th grid in the decomposition of \textsc{P3}. \\[2pt]
\IF{$d = 1$}
\STATE  return $\bv{u} = K_{\gridset_{\ell, 1}} \bv{v}$ \hfill \COMMENT{Base case}
\ENDIF\\[2pt]
\STATE \COMMENT{Pre-computation} \\
 \FOR{$i = 0$ to $\ell$} 
 \STATE ${\overline{A}}_i  = K_{\gridset_{i,1}} \mathcal{S}_{\gridset_{i,1}, \Omega_i} V_i $ \\
 \STATE   \change{  ${\overline{B}}_i^{T} = \textbf{mvm} \left(K_{\gridset_{\ell- i,d-1}},
  V_i^{T} \right) $ \hfill \COMMENT{Recursively multiply $K_{\gridset_{\ell- i,d-1}}$ with columns of $V_i^{T}$.} }\\
 \ENDFOR\\[2pt]
\STATE \COMMENT{Main loop}
 \FOR{$i = 0$ to $\ell$} 
  \STATE \change{$A_i^{T} = \textbf{mvm}  \left(  K_{\gridset_{\ell- i, d-1}}, 
    \left(\sum\limits_{j > i} \mathcal{S}_{\Omega_i, \gridset_{j,1}} {\overline{A}}_{j} 
    \mathcal{S}_{\gridset_{\ell-j,d-1},\gridset_{\ell-i, d-1}}\right)^{T} \right)$ \hfill \COMMENT{Recurse like line 8.} }\\
\STATE    $B_i = \mathcal{S}_{\Omega_i, \gridset_{i, 1}} K_{\gridset_{i,1}} 
 \left( \sum\limits_{j \leq i} \mathcal{S}_{\gridset_{i,1}, \Omega_j} {\overline{B}}_{j} \mathcal{S}_{\gridset_{\ell- j, d-1}, \gridset_{\ell - i, d-1}} \right)$ \\
\STATE    $\bv{u}_i = \vectorize \left[A_i \right]  +\vectorize \left[B_i \right]$
 \ENDFOR
\end{algorithmic}
\end{algorithm}
\end{mdframed}
\end{figure}

\begin{mdframed}[backgroundcolor=light-gray]

\begin{theorem}\label{thm:sparse_grid_mvm_algorithm}  
\change{
  Let $K_{\gridset_{\ell,d}}$ be the kernel matrix for 
  a $d$-dimensional sparse grid with resolution $\ell$ for 
  a stationary product kernel. For any $\bv v \in \mathbb{R}^{\cardinalitynorm{\gridset_{\ell,d}}}$, 
Algorithm~\ref{alg:sgkma_algorithm} computes $K_{\gridset_{\ell,d}}\bv v$ 
in $\O(\ell^{d} 2^\ell)$ time. 
}

\end{theorem}
\end{mdframed}

The formal analysis and proof of Theorem~\ref{thm:sparse_grid_mvm_algorithm} appears in 
Appendix $\ref{app:ski_on_sparse_grids}$. The running time of \change{$\O(\ell^{d} 2^\ell)$} is 
nearly linear in $\cardinalitynorm{\gridset_{\ell,d}}$ and 
much faster asymptotically than the naive MVM algorithm that materializes the full matrix and has running time quadratic in $\cardinalitynorm{\gridset_{\ell,d}}$.

Algorithm~\ref{alg:sgkma_algorithm} is built on two key observations.
First, the decomposition of Property~\textsc{P3} from Proposition~\ref{proposition:sparse_grid_properties}
\change{and the fact that the kernel follows product structure across dimensions are} used to decompose the MVM into blocks, each of which is between sub-grids
which are the product of a 1-dimensional rectilinear grid and a
sparse grid in $d-1$ dimensions.
Therefore, the overall MVM computation can be recursively decomposed into MVMs with
sparse grid kernel matrices in $d-1$ dimensions. This observation is in part inspired by \citet{fastmvmgalerkinmethod}, which also decomposes computation with matrices on sparse grids by the resolution of first dimension.
\change{The base case occurs when $d=1$. We assume that Toeplitz structure, which arises due to the kernel being stationary, is leveraged to perform this base case MVM in $O(\ell 2^\ell)$ time. 
Algorithm~\ref{alg:sgkma_algorithm} can also be extended to \emph{non-stationary} product kernels by using a standard MVM routine for the base case, which changes the overall running-time analysis but is still more efficient than the naive algorithm.}

Secondly, by Property~\textsc{P2}, the kernel matrix multiplication for any grid of resolution $\ell$
also includes the result of the multiplication for grids of lower resolution and the same number of dimensions. 
Thus, the results of the multiplications for many individual blocks 
can be obtained by using the appropriate selection operators 
with the result of the multiplication with the kernel
 matrix $K_{\gridset_{\ell-i, d-1}}$ in Line~12 of the algorithm. 
Further intuition and explanation are provided in the Appendix \ref{app:fast_mvm_algorithm}.

\medskip

\noindent \textbf{Improving batching efficiency.} 
The recursions in Lines 8 and 12 can be batched for efficiency, since both are multiplications with the
same symmetric kernel matrix $K_{\gridset_{\ell, d}}$.
Similarly, the recursion spawns many recursive multiplications with kernel matrices of the form $\gridset_{\ell', d'}$ for $0 \leq \ell' < \ell$ and $1 \leq d' < d$, and the calculation can be reorganized to batch all multiplications with each $K_{\gridset_{\ell', d'}}$. This is a significant savings, because there are only $d(\ell+1)$ distinct kernel matrices, but the recursion has a branching factor of $(\ell+1)$, so spawns many recursive calls with the same kernel matrices.

\subsection{Sparse Interpolation For Sparse Grids}
\label{ssec:interpolation_and_gaussian_processes}

We now seek to construct the matrix $W$, which interpolates function values from the sparse grid $\gridset_{\ell, d}$ to training points $\x_i \in \R^d$, while ensuring that each row of $W$ is sparse enough to preserve efficiency of matrix-vector multiplications with $W$. This requires a sparse interpolation rule for sparse grids.

To set up the problem, we consider interpolating a function $f$ observed at points in a generic set $U$. Let $\bv f_U = (f(\x))_{u \in U}$.
A linear interpolation rule for $U$ is a mapping $\bv x \mapsto \bv w_{\bv x} \in \R^{|U|}$ used to approximate $f(\x)$ at an arbitrary point as $f(\x)\approx \bv w_{\bv x}^T \bv f_U$.
The density of an interpolation rule is the maximum number of non-zeros in $\bv w_{\bv x}$ for any $\bv x$.

The \emph{combination technique} for sparse grids constructs an interpolation rule by combining interpolation rules for the constituent rectilinear grids.

\begin{mdframed}[backgroundcolor=light-gray]
\begin{proposition}\label{proposition:sparse_grid_interpolation}
For each $\bv l$, let $\bv w_{\bv x}^{{\bv l}}$ be an interpolation rule for the rectilinear grid $\Omega_{\bv l}$ with maximum density $c$. The combination technique gives an interpolation rule $\bv w_\bv x$ for the sparse grid $\gridset_{\ell, d}$ with density at most $c \times \binom{\ell + d -1}{d-1}$. The factor $\binom{\ell + d -1}{d-1}$ is the number of rectilinear grids included in $\gridset_{\ell, d}$.
\end{proposition}
\end{mdframed}

We use the combination technique to construct the sparse-grid interpolation coefficients $\bv w_{\bv x_i}$ for each training point $\bv x_i$ and stack them in the rows of $W$, which can be done in time proportional to the number of nonzeros. 
Details are given in Appendix~\ref{app:background}. The combination technique can use any base interpolation rule for the rectilinear grids, such as multilinear, cubic, or simplicial interpolation.

\subsection{Simplicial Interpolation}

\begin{table}[h]
  \centering
  \caption{
  \change{    
  The MVM complexities (i.e., $\mv(\cdot)$) of the interpolation matrix $W$ and the kernel matrix $K_G$
  for different interpolation bases and $d$-dimensional grids 
  with $2^{\ell}$ unique points in each dimension and $n$ data points. 
  A star indicates approaches proposed in this work. 
  `Dense' denotes a rectilinear grid. 
  }
  }
  \label{table:mvm_complexity}
  \vspace*{2mm}
  \begin{tabular}{|l|c|c|c|}
  \hline
  Grid                       & Basis &  $\mv(W)$ &  $\mv(K_G)$  \\ \hline
  \multirow{3}{*}{Dense} & Cubic          &    $\O(n \cdot 4^d)$    &  $\O(\ell \cdot d \cdot 2^{\ell \cdot d})$     \\ \cline{2-4} 
                                  & Linear         &     $\O(n \cdot 2^d)$   &   $\O(\ell \cdot d \cdot 2^{\ell \cdot d})$     \\ \cline{2-4} 
                                  & Simplicial$^{\star}$   &    $\O(n \cdot d^2)$    &   $\O(\ell \cdot d \cdot 2^{\ell \cdot d})$      \\ \hline
  \multirow{3}{*}{Sparse}    & Cubic$^{\star}$         &     $\O(n \cdot 4^d \cdot  \binom{\ell+d-1}{d-1})$     &     $\O(\ell^d \cdot 2^{\ell})$     \\ \cline{2-4} 
                                  & Linear$^{\star}$        &    $\O(n \cdot 2^d \cdot  \binom{\ell+d-1}{d-1})$      &  $\O(\ell^d \cdot 2^{\ell})$       \\ \cline{2-4} 
                                & Simplicial$^{\star}$     &    $\O(n \cdot d^2 \cdot  \binom{\ell+d-1}{d-1})$      &    $\O(\ell^d \cdot 2^{\ell})$     \\ \hline
\end{tabular}
\end{table}

The density of the interpolation rule is a critical consideration for kernel interpolation techniques -- with or without sparse grids.
Linear and cubic interpolation in $d$ dimensions have density $\Theta(2^d)$ and $\Theta(4^d)$, respectively.
This ``second curse of dimensionality'' makes computations with $W$ intractable in higher dimensions independently of operations with the grid kernel matrix.
For sparse grids, the density of $W$ increases by an additional factor of $\binom{\ell + d -1}{d-1}$.

We propose to use simplicial interpolation~\cite{halton1991simplicial} for the underlying interpolation rule to avoid exponential growth of the density.
Simplicial interpolation refers to a scheme where $\R^d$ is partitioned into simplices and a point $\x$ is interpolated using only the $d+1$ extreme points of the enclosing simplex, so the density of the interpolation rule is exactly $d+1$.
Simplicial interpolation was previously proposed for sparse grid classifiers in~\cite{garcke2002classification}.
In work closely related to ours, \citet{SimplicesGP} used simplicial interpolation for GP kernel interpolation, with the key difference that they use the \emph{permutohedral lattice} as the underlying grid, which has a number of nice properties but does not come equipped with fast specialized routines for kernel matrix multiplication.

In contrast to \citet{SimplicesGP}, we maintain rectilinear and/or sparse underlying grids, which preserve structure that enables fast kernel matrix multiplication.
For rectilinear grids, this requires partitioning each hyper-rectangle into simplices, so the entire space is partitioned by simplices whose extreme points belong to the rectilinear grid.
Then, within each simplex, the values at the extreme points are interpolated linearly.
In general, there are different ways to partition hyper-rectangles into simplices -- we use the specific scheme detailed in~\cite{halton1991simplicial}.
For sparse grids, we then use the combination technique, leading to overall density of $(d+1)\binom{\ell+d-1}{d-1}$. 
\change{Table \ref{table:mvm_complexity} provides the MVM 
complexities of $W$ and kernel matrices for 
different interpolation schemes and both grids.}
\change{More details on how to perform simplicial interpolation with rectilinear and sparse grids}
are given in the
Appendix~\ref{app:simplicial_interpolation_onrectlinear_grids}.

\section{Experiments}
\label{sec:experiments}
In this section, we empirically evaluate the time and memory taken by Algorithm 1 for matrix-vector multiplication with the sparse grid kernel matrix, 
the accuracy of sparse grid interpolation and GP regression as the data dimension $d$ increases,
and the accuracy of sparse grid kernel interpolation for GP regression on real higher-dimensional datasets from UCI.
Hyper-parameters, data processing steps, and optimization details are given in Appendix \ref{app:hyperparameters_details}.

\begin{figure}[h]
    \begin{subfigure}{0.48 \textwidth}
    \centering \includegraphics[width=0.9\textwidth]{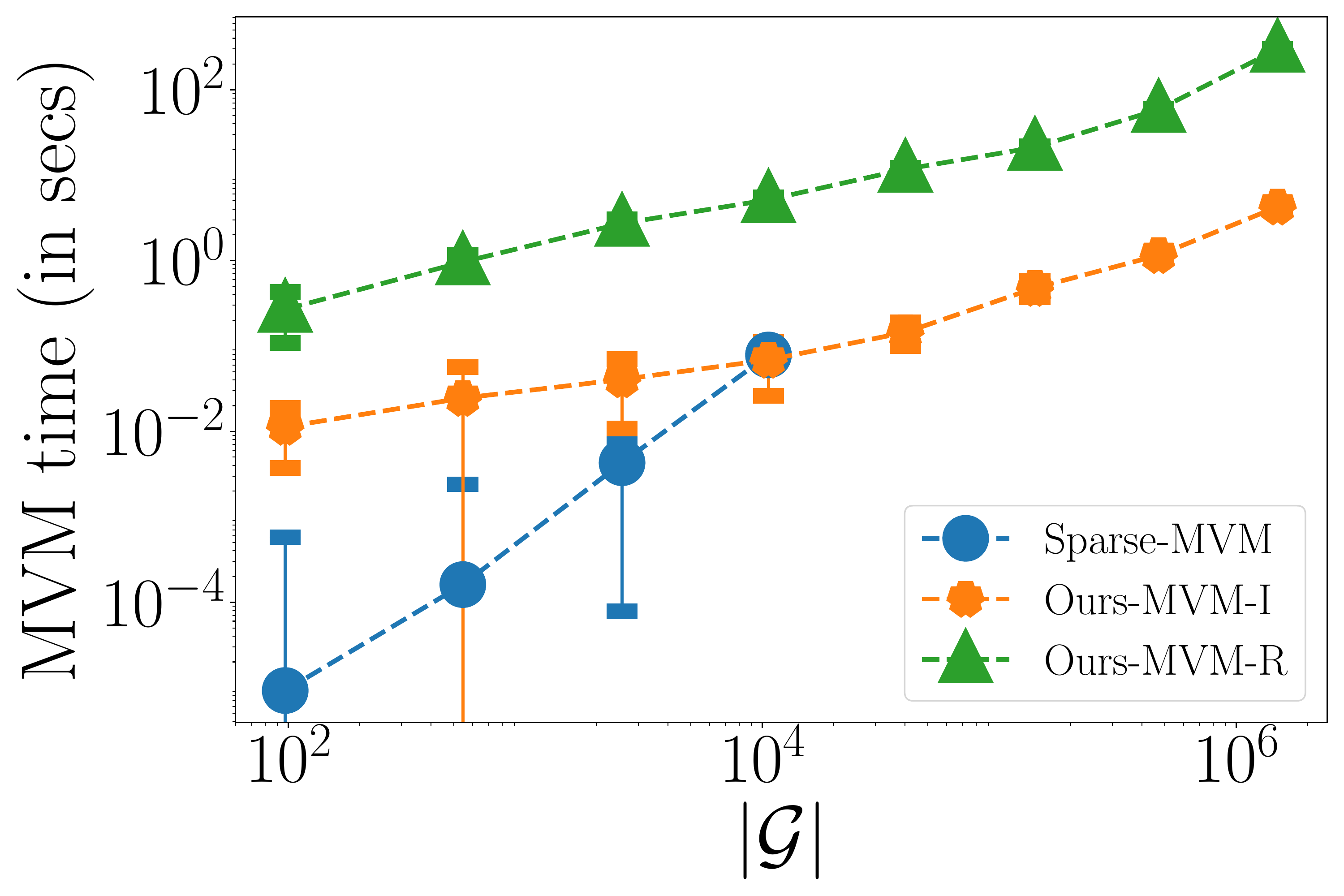}
    \end{subfigure}
    \hfill
    \begin{subfigure}{0.48 \textwidth}
    \centering \includegraphics[width=0.9\textwidth]{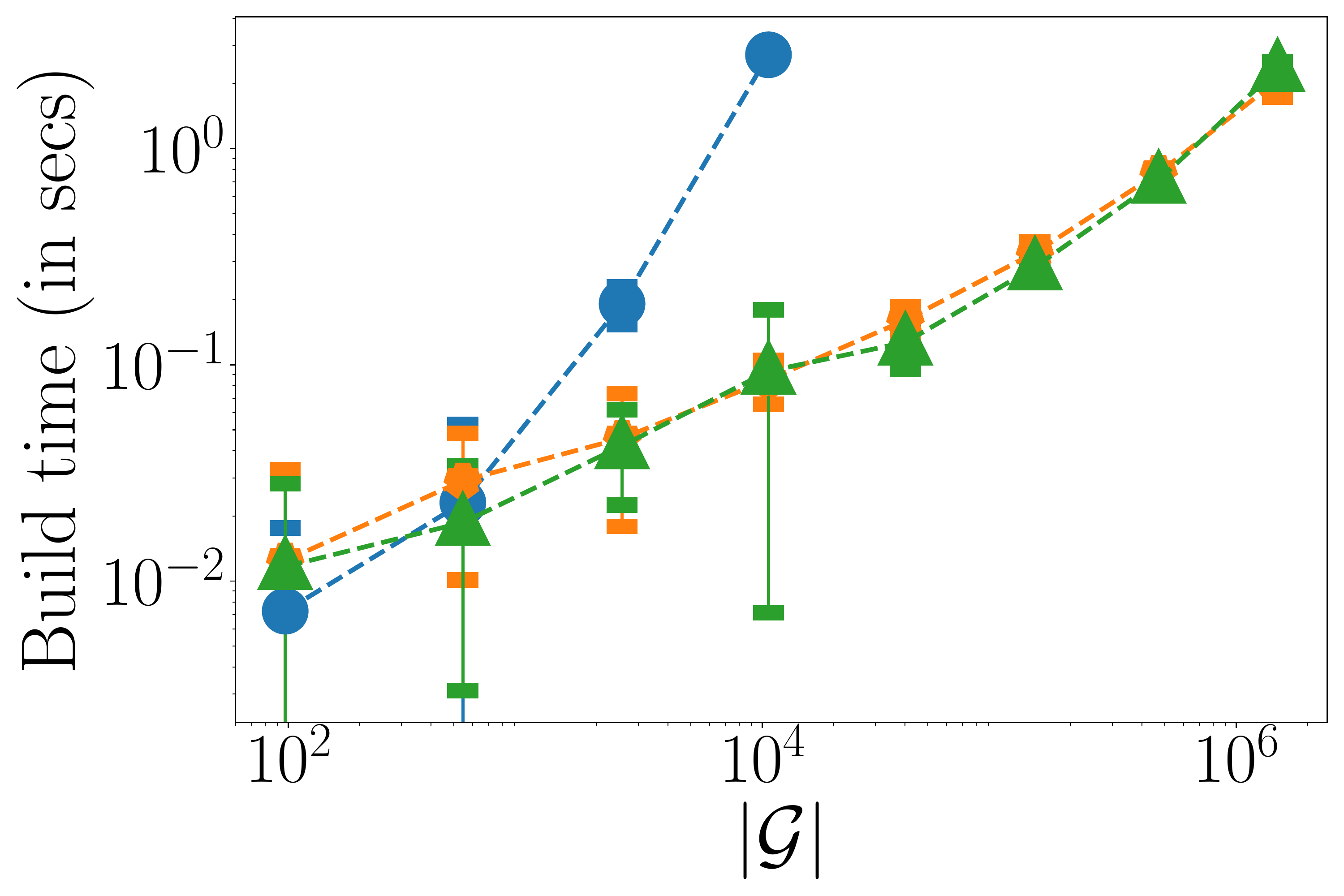}
    \end{subfigure}
    \hfill
    \begin{subfigure}{0.48 \textwidth}
    \centering \includegraphics[width=0.9\textwidth]{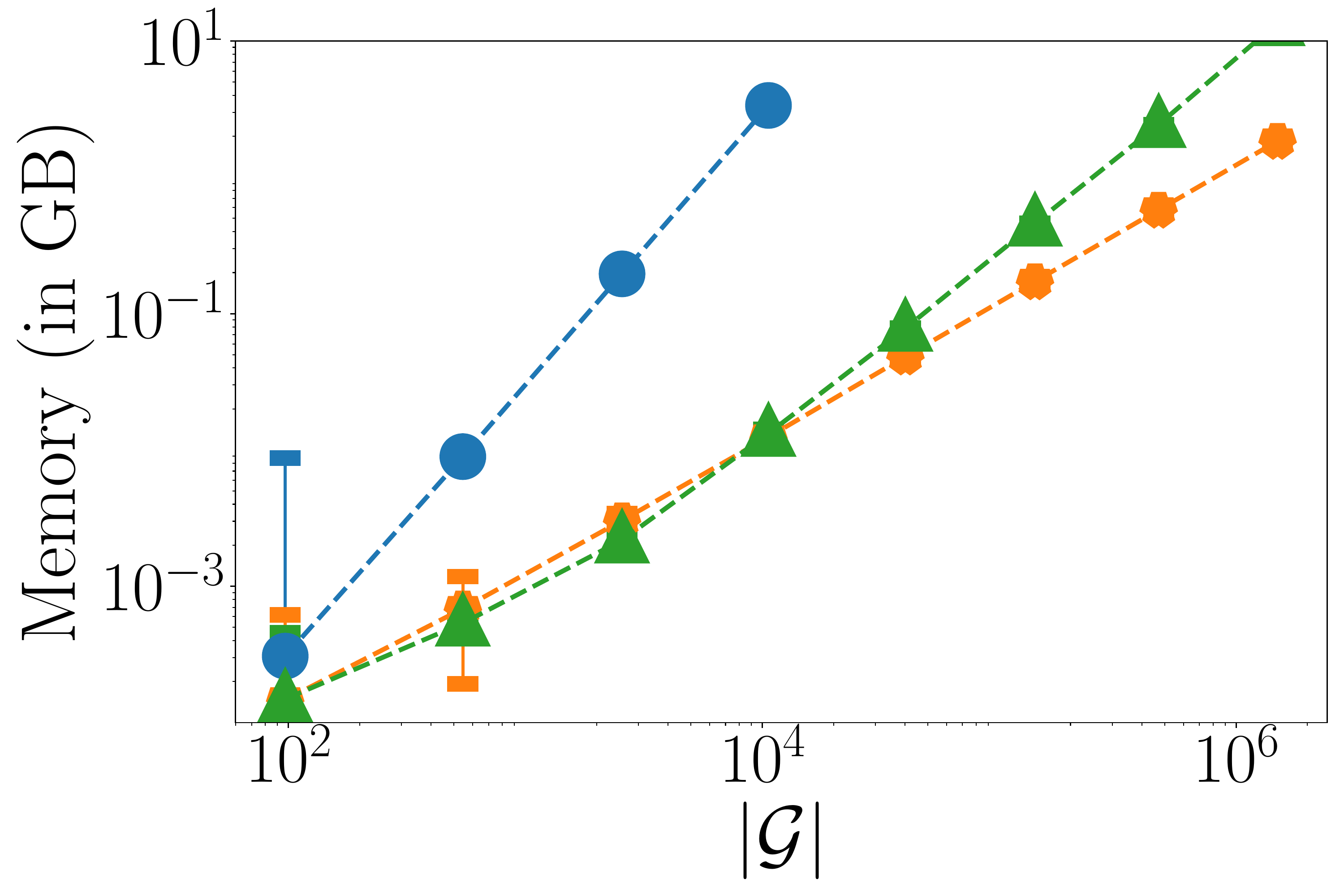} 
    \end{subfigure}
    \hfill
    \begin{subfigure}{0.48 \textwidth}
    \centering \includegraphics[width=0.9\textwidth]{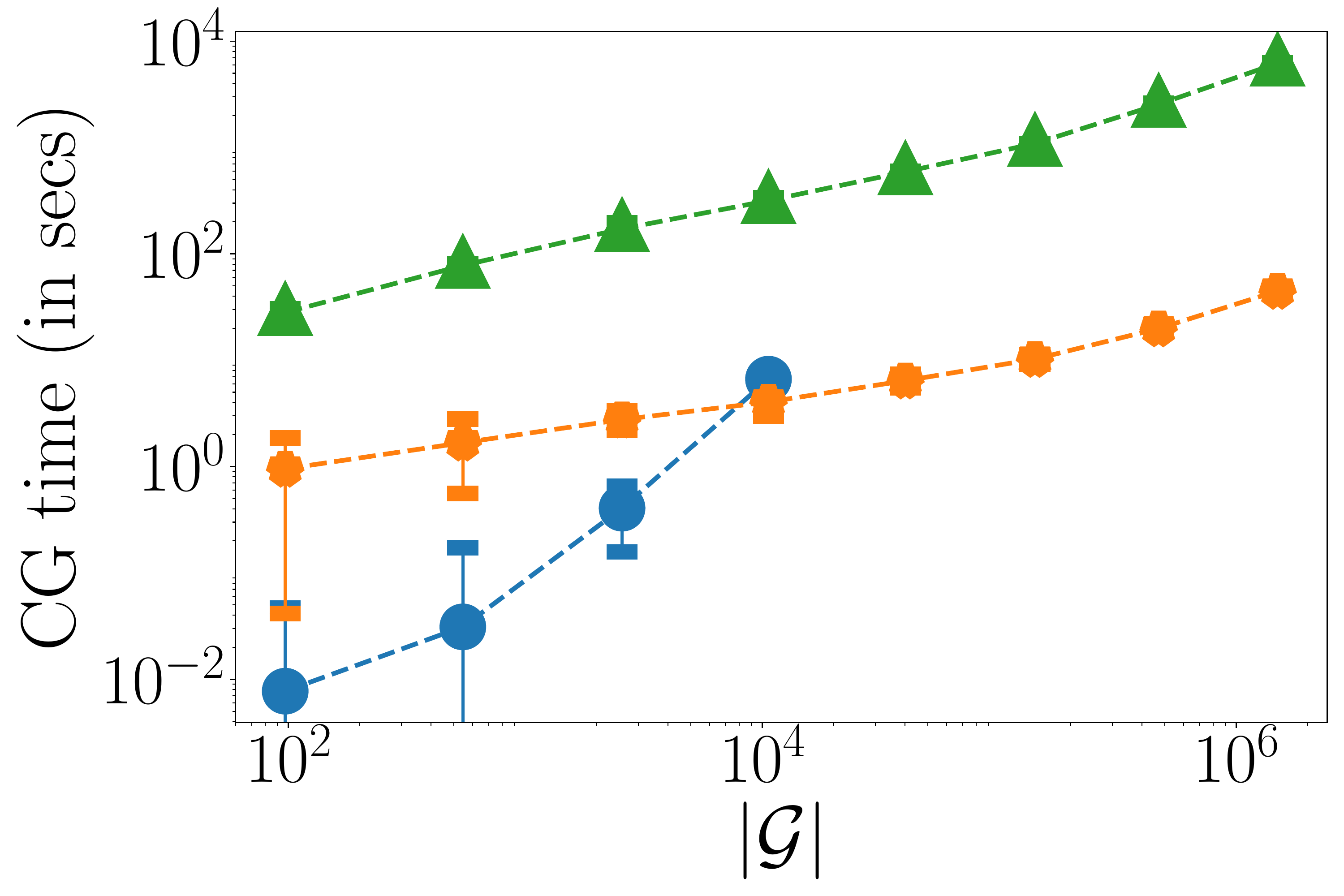} 
    \end{subfigure}
    \caption{Matrix-vector multiplication (MVM) resource usage relative to sparse grid 
    size for $d=6$ and increasing resolution $\ell$. From top-left to bottom-right: 
    time for one MVM operation, build (pre-processing) time for the kernel matrix,
    peak memory usage, and typical time to solve a linear system 
    using CG (time for build plus $50$~MVMs). Each plot shows
    the naive quadratic MVM algorithm (Sparse-MVM), the recursive implementation of Algorithm~\ref*{alg:sgkma_algorithm} (Ours-MVM-R), and the efficiently batched iterative implementation of Algorithm~\ref*{alg:sgkma_algorithm} (Ours-MVM-I). All measurements are averaged over $8$ trials; error bars represent twice the standard error.} 
\label{fig:mvm_time_memory_comparison}
\end{figure}

\medskip

\noindent \textbf{Sparse grid kernel MVM complexity.} 
First, we evaluate the efficiency of MVM
algorithms. We compare the basic and efficient implementations of Algorithm~\ref{alg:sgkma_algorithm}
to the naive algorithm, which constructs the full kernel matrix and scales quadratically
with the sparse grid size. 
Algorithm~\ref*{alg:sgkma_algorithm} has a significant
theoretical advantage in terms of both time and memory requirements as the grid size grows.
Figure \ref{fig:mvm_time_memory_comparison} illustrates this for $d=6$ by comparing
MVM time and memory requirements for sparse grids with resolutions $\ell \in \{2, \ldots, 9\}$ (roughly 100 to 1M grid points).
The MVM time, preprocessing time, and memory consumption of Algorithm~\ref{alg:sgkma_algorithm} all grow more slowly than the naive algorithm, and the efficient implementation of Algorithm~\ref{alg:sgkma_algorithm} is faster for $\cardinalitynorm{\gridset}$ larger than about $10^4$, after which the naive algorithm also exceeds the 10~GB memory limit.
For comparison, at $\ell=6$ (about 40K grid points), Algorithm~\ref{alg:sgkma_algorithm} uses only $0.05$~GB of memory.
These results indicate that the proposed algorithm is crucial for enabling sparse grid kernel interpolation in higher dimensions.
Figure \ref{fig:mvm_time_memory_comparison} also depicts the typical time to run GP inference (i.e., preprocessing time plus $50$ MVM operations).

\begin{figure}[h]
    \begin{subfigure}{0.45 \textwidth}
    \centering \includegraphics[height=40mm]{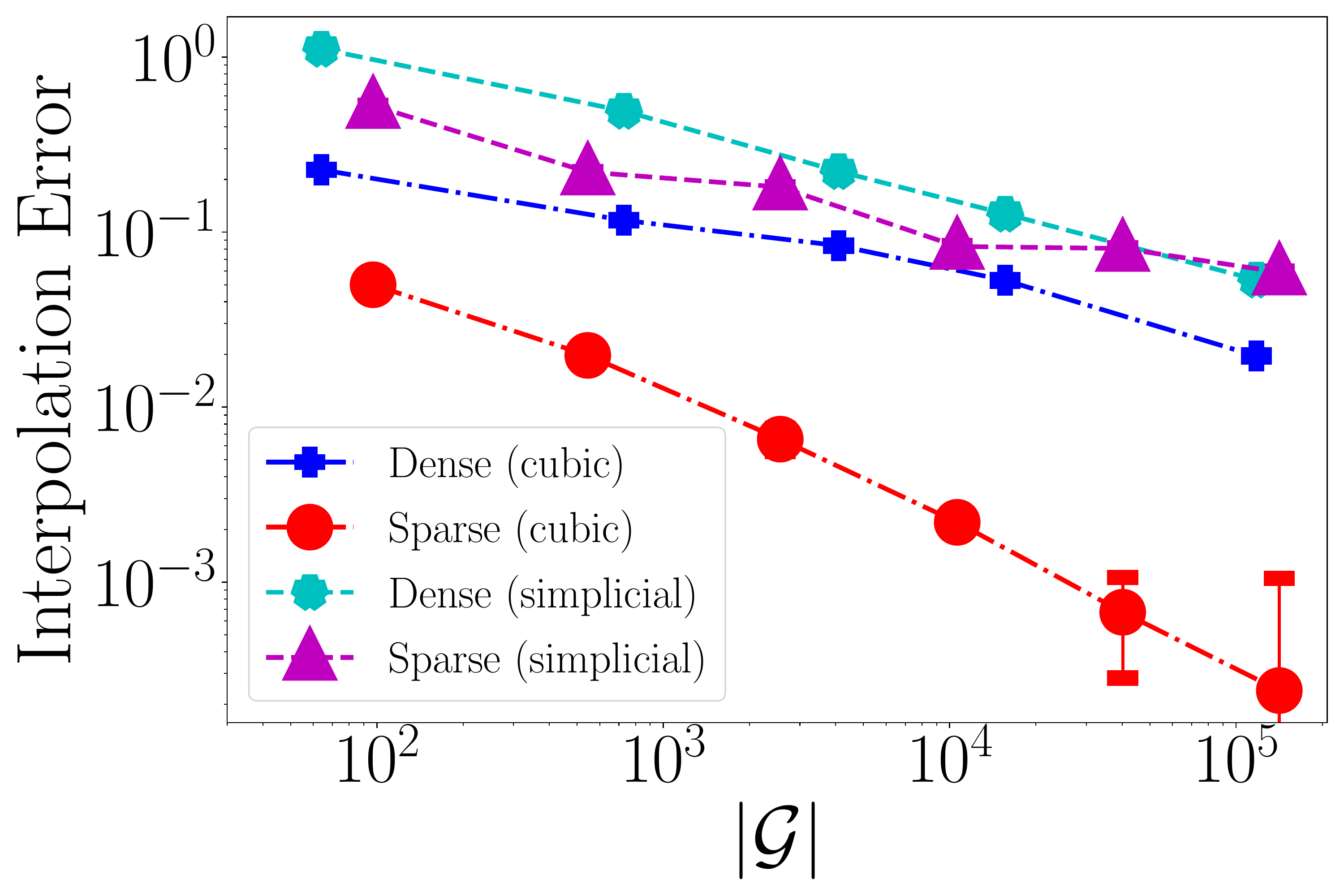}
    \end{subfigure}
    \hfill
    \begin{subfigure}{0.45 \textwidth}
    \centering \includegraphics[height=40mm]{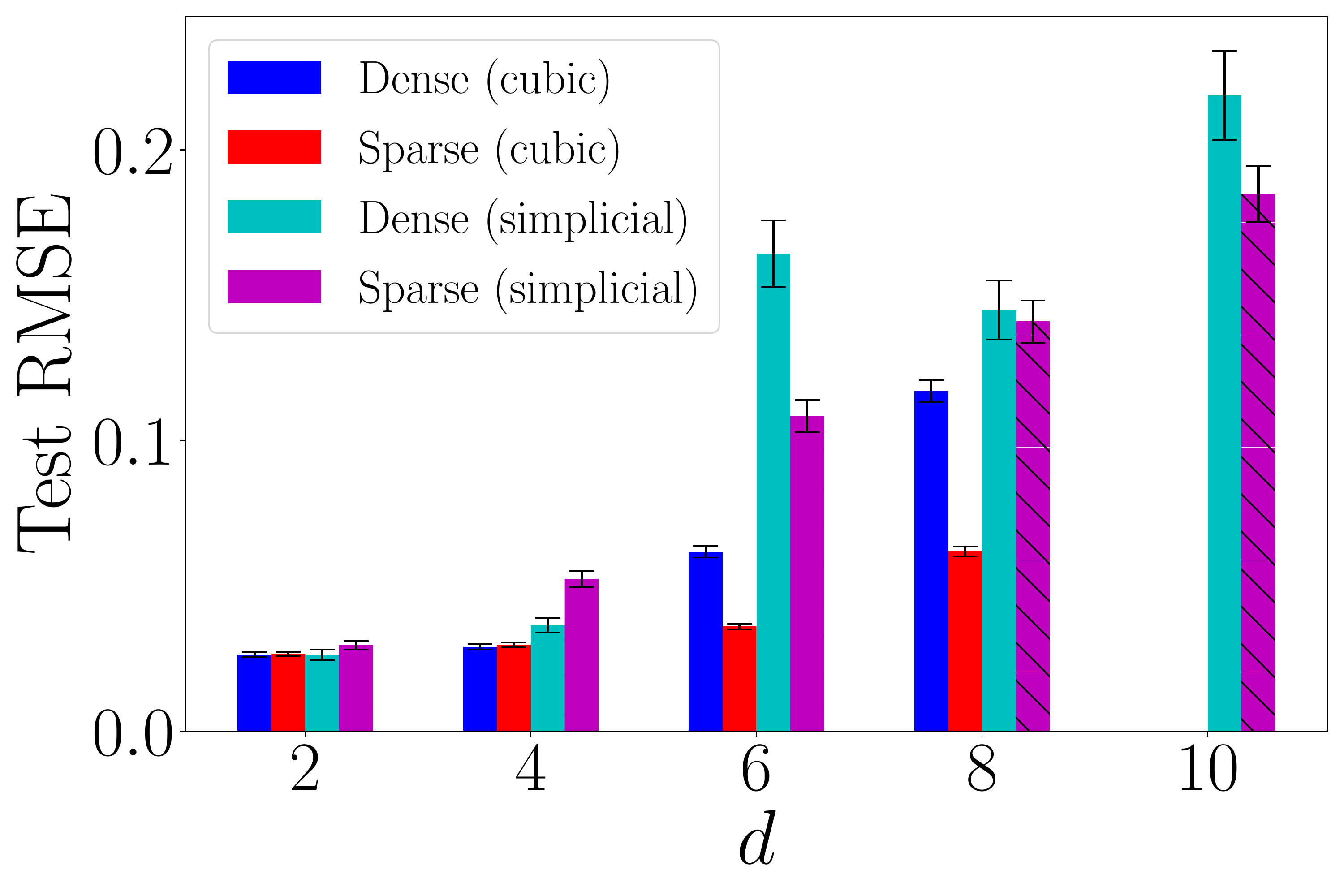} 
    \end{subfigure}
    \caption{Comparing dense and sparse grids with cubic and simplicial 
    interpolation schemes on synthetic data. 
    Dense (cubic/simplicial) are dense grid methods 
    with cubic/simplicial interpolation schemes; similarly, 
    sparse (cubic/simplicial) are for sparse grids.
    Left: Function interpolation error vs. the grid size for $d=6$.  
    Right: Test root-mean-square error (RMSE) for GP regression for increasing dimensions.
    For both tasks, sparse grid methods outperform dense grids. For $d=10$,
    both methods with cubic interpolation run out of GPU memory, which is $48$ GB for this experiment.
    }
\label{fig:function_interpolation_gp_inference_results}
\end{figure}

\medskip

\noindent\textbf{Sparse grid interpolation and GP inference accuracy on synthetic data.} 
A significant advantage of sparse grids over dense rectilinear grids is their ability 
to perform accurate interpolation in higher dimensions.
We demonstrate this for $d=6$ by interpolating
the function $f(\bv{x}) = \cos(\norm{\bv{x}}_1)$ from observation locations on sparse grids of increasing
resolution onto 200 random points sampled uniformly from $[0, 1]^d$. 
Figure~\ref{fig:function_interpolation_gp_inference_results}, left, shows the interpolation
error for dense and sparse grids with both cubic and simplicial interpolation.
\textit{Sparse (cubic)} is significantly more accurate than \textit{Dense (cubic)}, and \textit{Sparse (simplicial)} is more accurate than \textit{Dense (simplicial)}.

We next evaluate the accuracy of GP inference in increasing dimensions. 
We keep the same function $f(\bv{x})$ and generate observations
as $y = f(\bv{x}) + \mathcal{N} (0, 0.05)$.
For all $d$, we use $\ell = 4$ for the sparse grid and compare to the
dense grid with the closest possible total number of grid 
points
(i.e., $\lceil \gridset_{4, d}^{1/d} \rceil$ points in each 
dimension)
We tried to match the sizes of both dense and sparse grids while ensuring that the dense grid always had at least as many points as the sparse grid, to give a fair comparison. Precisely, (d, dense grid size, sparse grid size) tuples are
$(2, 144, 129) , (4, 1296, 796), (6, 4096, 2561), (8, 6561, 6401), (10, 59049, 13441)$. 
For $d \geq 8$, performance is better with sparse grids than dense grids for 
both interpolation schemes.
Remarkably, our proposal to use simplicial interpolation with dense grids allows SKI to scale to $d=10$, 
which is a significant improvement over prior work, 
in which SKI is typically infeasible for $d \geq 4$.

\begin{table}[h]
\caption{Test root-mean-square-error (RMSE) on UCI regression datasets 
with dimensions (i.e., $8 \leq d \leq 10$).
See text for algorithm descriptions and settings.
All mean and standard deviations are computed over three trials.
}
\vspace*{2mm}
\centering
\small{
\begin{tabular}{|l|c|c|c|c|c|}
\hline
Datasets ($d$) & SGPR & SKIP & Simplex-GP & Dense-grid & Sparse-grid \\ \hline
\textit{Energy} ($8$) & $1.509\pm 0.003$ & $5.762 \pm 0.000$ & $3.076\pm 0.012$ & $1.333\pm 0.009$ & $\mathbf{0.715} \pm 0.004$ \\ \hline
\textit{Concrete}  ($8$) & $12.727 \pm 0.018$ & $12.727 \pm 0.001$ &  $12.727\pm 0.000$  &  $12.191 \pm 0.001$  & $\mathbf{8.655}\pm 0.002$ \\ \hline
\textit{Kin40k}   ($8$) &  $\mathbf{0.168} \pm 0.009$ & $0.174 \pm 0.001$ & $0.287 \pm 0.003$     & $0.205 \pm 0.003$    &  $0.483 \pm 0.000$\\ \hline        
\textit{Fertility}  ($9$) & $0.197 \pm 0.016$ & $0.183 \pm 0.000$ &  $0.187 \pm 0.001$ & $\mathbf{0.182} \pm 0.000$ & $0.194 \pm 0.002$ \\ \hline
\textit{Pendulum} ($9$) & $\mathbf{1.948} \pm 0.021$ & $2.947 \pm 0.000$ &  $2.577 \pm 0.009$ & $2.053\pm 0.010$ & $2.103\pm 0.015$ \\ \hline
\textit{Protein}   ($9$) & $0.605 \pm 0.001$ & $0.778 \pm 0.000$ & $\mathbf{0.582} \pm 0.018$    & $0.736 \pm 0.002$   &  $0.595 \pm 0.001$\\ \hline   
\textit{Solar}  ($10$) & $0.790 \pm 0.026$ & $0.780 \pm 0.002$& $ 0.792 \pm 0.000$ & $0.775 \pm 0.002$ & $\mathbf{0.748} \pm 0.006$ \\
\hline
\end{tabular}
}
\label{table:rmse_on_uci_datasets}
\end{table}

\paragraph{GP regression performance on UCI datasets.} 
To evaluate the effectiveness of our methods for scaling GP kernel interpolation to higher dimensions, we consider all UCI \citep{ucirepo} data sets with dimension $8 \leq d \leq 10$.
We compare our proposed methods, Dense-grid (dense SKI with simplicial interpolation) and Sparse-grid (sparse-grid SKI with simplicial interpolation), to SGPR~\citep{titsias2009variational}, SKIP~\citep{gardner2018product}, and Simplex-GP~\citep{SimplicesGP}.
For SGPR, we report the best results using $256$ or $512$ inducing points. 
For SKIP, $100$ points per dimension are used. 
For Simplex-GP \citep{SimplicesGP}, the blur stencil order is set to $1$.
Table \ref{table:rmse_on_uci_datasets} shows the root mean squared error (RMSE) for all methods.
Our methods have performance comparable to and often better than SGPR and prior
methods for scaling kernel interpolation to higher dimensions.
This shows that sparse grids and simplicial interpolation can effectively scale
SKI to higher dimensions to give a GP regression framework that is competitive with
state-of-the-art approaches.
We report additional results and analysis in Appendix~\ref{app:experiments}.

\section{Related Works}

Beyond SKI and its variants, a number scalable GP approximations have been investigated.
Most notable are the different variants of sparse GP approximations~\citep{williams2001using,Snelson2005SparseGP,quinonero2005unifying}. For $m$ inducing points, these methods require either $\Omega (nm^2)$ time for direct solves or $\Omega(nm)$ time for approximate kernel MVMs in iterative solvers. While these methods generally do not leverage structured matrix algebra like the SKI framework, and thus have worse scaling in terms of the number of inducing points $m$, they may achieve comparable accuracy with smaller $m$, especially in higher dimensions. Some work utilizes the SKI framework  to further boost the performance of sparse GP approximations~\citep{PavelBillionOfInducingPoints}. 

A closely related work to ours is on improving the 
dimension scaling of SKI through the use of low-rank approximation and product structure~\citep{gardner2018product}.  
In another very closely related work, \citet{SimplicesGP} recently proposed to scale SKI to higher dimensions
via interpolation on a permutohedral lattice. 
Like our work, they use simplicial interpolation.
Unlike our work, the kernel matrix on the permutohedral lattice does not have special structure that admits fast exact multiplications; they instead use a locality-based approximation that takes into account the length scale of the kernel function by only considering pairs of grid points within a certain distance.

Another related direction of research 
is the adaptation of sparse grid techniques for machine learning problems, 
e.g., classification and regression \citep{pfluger2010spatiallyclassification} 
and data mining \citep{bungartz2008adaptivedatamining}.
These methods construct feature representations using sparse grid points~\citep{tridao}, often
by selecting a subset of grid points 
\citep{garcke2006dimensioncombinationtechnqiquemachinelearning,tridao}.

Similar to this work, \citet{plumlee2014fast} proposed the use of sparse grid with GPs. Their 
work primarily focuses on the experimental design problem, 
which permits the
observation locations to be only on the sparse grids. 
In contrast, the GP regression problem necessitates interpolation, as the observations are not required to be on the grid.  Methodologically, 
\citet{plumlee2014fast} employs 
a direct approach to invert the sparse grid kernel matrix, as opposed to the MVMs used within the SKI framework and this work. 
Additionally, various methods have been proposed to adapt sparse grids for higher dimensions. These methods utilize a subset of rectilinear grids and apply differential scaling across dimensions \cite{saad1986gmres, pfluger2010spatiallyclassification, plumlee2021composite}. 




\section{Discussion}
\label{sec:discussion}

\change{
    This work demonstrates that two classic numerical techniques, namely, 
    sparse grids and simplicial interpolation, 
    can be used to scale GP kernel interpolation to higher dimensions. 
    SKI with sparse grids and simplicial interpolation has 
    better or competitive regression accuracy compared to state-of-the-art GP regression approaches on several UCI benchmarking datasets with 8 to 10 dimensions.   
}

\medskip

\noindent\textbf{Limitations and future work.}
Sparse grids and simplicial interpolation address two important bottlenecks when scaling kernel interpolation to higher dimensions. Sparse grids allow scalable matrix-vector multiplications with the grid kernel matrix, and simplicial interpolation allows scalable multiplications with the interpolation matrix $W$. 
The relatively large number of rectilinear grids used to form a sparse grid -- i.e., the factor of $\binom{\ell+d-1}{d-1}$ in Proposition~2 -- is one limiting factor that makes multiplication by $W$ more costly. 
Future research could investigate methods to mitigate this extra cost, and explore the limits of scaling to even higher dimensions with sparse grids.

\bibliographystyle{plainnat}
\bibliography{main} 

\clearpage
\appendix
\onecolumn

\begin{appendices}
\begin{center} \Large \textbf{Supplementary Appendices}
\end{center}

\section{Background -- Omitted details}\label{app:background}

\subsection{Sparse grids - Visualizations of grid points}
\begin{figure}[h]
    \centering \includegraphics[width=0.8\textwidth, height=70mm]{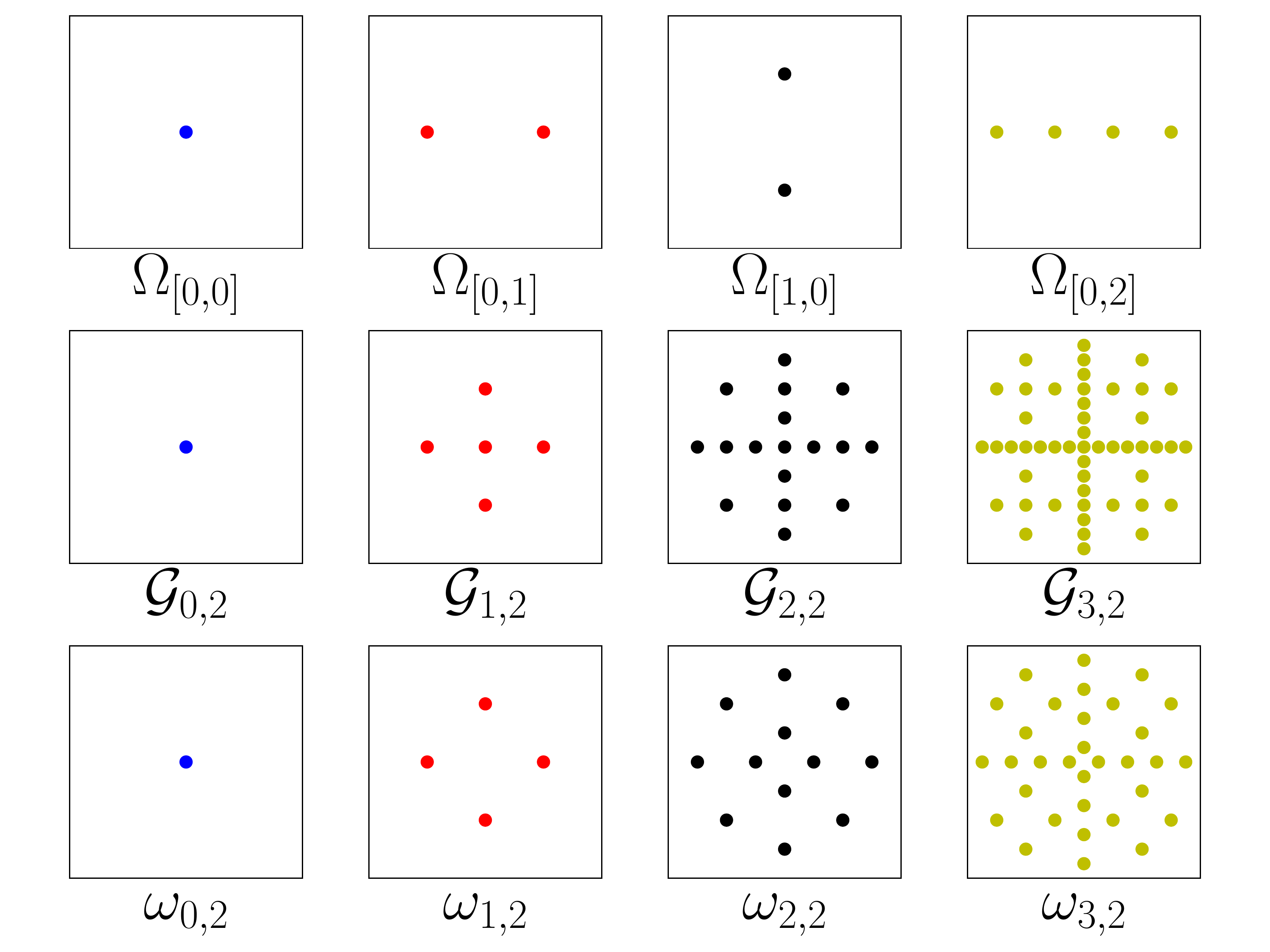}
    \caption{Visualizations of grid points on $\left[0, 1 \right] \times \left[0, 1 \right]$. 
    From top to bottom: (row-1) rectilinear grids with different resolution vectors, 
    (row-2) sparse grids with different resolutions for $d=2$, and (row-3) incremental gain of grid points 
     with the resolution, i.e., $\omega_{\ell, d} \definedas \gridset_{\ell, d}  \setminus \bigcup_{0 \leq  l' \leq \ell-1} \gridset_{l', d}$.
    }
\label{fig:sparse_grid_locatgions}
\end{figure}


\subsection{Sparse grids -- Properties and Hierarchical Interpolation}
\label{app:sparse_grids}

\begin{repproposition}
    {proposition:sparse_grid_properties} [\textbf{Properties of Sparse Grid}]
    Let $\gridset_{\ell, d} \subset [0, 1]^d$ be a sparse grid with any resolution $\ell \in \mathbb{N}_{0}$ and dimension $d\in \mathbb{N}$. Then the following properties hold:
\begin{align*}
    &\textsc{(P1)} \,\,\,  \cardinalitynorm{\gridset_{\ell}^{d}} = \O(2^\ell \ell^{d-1}), \\
    &\textsc{(P2)} \,\,\, \forall \ell^{'} \in \mathbb{N},  0 \leq \ell^{'} \leq \ell \,\,\, \implies  \gridset_{\ell^{'},d} \subseteq \gridset_{\ell, d}, \\
    &\textsc{(P3)} 
        \,\,\, \gridset_{\ell, d} = \bigcup {}_{i=0}^{\ell} \big(\Omega_i \otimes \gridset_{\ell-i, d-1}\big) \text{ and } \,\,\, \gridset_{\ell, 1} = \bigcup {}_{i=0}^{\ell} \Omega_{i}.
\end{align*}
\end{repproposition}
\begin{proof}
$\newline$ \textsc{(P1)} -- size of sparse grids:
\begin{align*}
    \cardinalitynorm{\gridset_{\ell}^{d}} &= \sum_{ \{ \bv{l} \in \mathbb{N}_0^{d} \mid \norm{\bv l}_1 \leq \ell  \} } (2\times 2^{\bv{l}_1-1}) (2\times 2^{\bv{l}_2-1}) \cdots (2\times 2^{\bv{l}_d-1}) \\
    &= \sum_{ \{ \bv{l} \in \mathbb{N}_0^{d} \mid \norm{\bv l}_1 = \ell  \} } 2^{\norm{\bv{l}}_1} \underbrace{\leq}_{\text{summing over resolution}} \sum_{0 \leq l' \leq \ell} \underbrace{{{l' + d -1} \choose {d-1}}}_{\leq {l' +  d -1 }^{d-1}}   2^{l'} \\
    &\leq (\ell + d -1 )^{d-1}  \sum_{0 \leq l' \leq \ell}   2^{l'} = \O(\ell^{d-1} 2^{\ell}) 
\end{align*}

$\newline$ \textsc{(P2)} -- sparse grids with the smaller resolution are in sparse grids with higher resolution:
\begin{align*}
\forall \ell^{'} \in \mathbb{N},  0 \leq \ell^{'} \leq \ell, \,\,\, \{ \bv{l} \in \mathbb{N}_0^{d} \mid \norm{\bv l}_1 \leq \ell^{'}  \} \subseteq \{ \bv{l} \in \mathbb{N}_0^{d} \mid \norm{\bv l}_1 \leq \ell  \} \implies  \gridset_{\ell^{'},d} \subseteq \gridset_{\ell, d}
\end{align*}
$\newline$ \textsc{(P3)} -- the recursive construction of sparse grids from rectilinear grids: \\ 

Let $\mathcal{L}_{\ell, d}$ be the set of $d$-dimensional vectors with $L_1$ norm bounded by $\ell$, i.e, 
$\mathcal{L}_{\ell, d} \definedas \{ \bv{l} \in \mathbb{N}_0^{d} \mid \norm{\bv l}_1 \leq \ell \}$. 
Notice that  $\mathcal{L}_{\ell, d}$ satisfies recursion similar to \textsc{P3}. 
I.e., $\mathcal{L}_{\ell, d} = \bigcup {}_{i=0}^{\ell} \big(\{i \} \otimes \mathcal{L}_{\ell-i, d-1}\big) \text{ and } \,\,\, \mathcal{L}_{\ell, 1} = \bigcup {}_{i=0}^{\ell} \{i \}$. 
Next, \textsc{P3} follows from the fact that $ \gridset_{\ell,d} =  \{ \Omega_{\bv l}  \mid \bv{l} \in \mathcal{L}_{\ell, d} \}$.
\end{proof}

\subsubsection{Sparse grids - A hierarchical surplus linear interpolation approach}
\label{sssec:hierar_sparse_interp}

This subsection demonstrates how to use hierarchical surplus linear interpolation for kernel interpolation with sparse grids. However, we do not explore this method in our experiments.
Nevertheless, we believe that the steps taken in adopting this method to kernel interpolation might be of interest to readers and plausibly helpful for future exploration of kernel interpolation with sparse grids.

Our exposition to sparse grids in section \ref{app:background} has been limited to 
specifying grid points, which can be extended to interpolation by associating basis functions with grid points.

\begin{figure}[h]
    \centering \includegraphics[width=0.8\textwidth] 
    {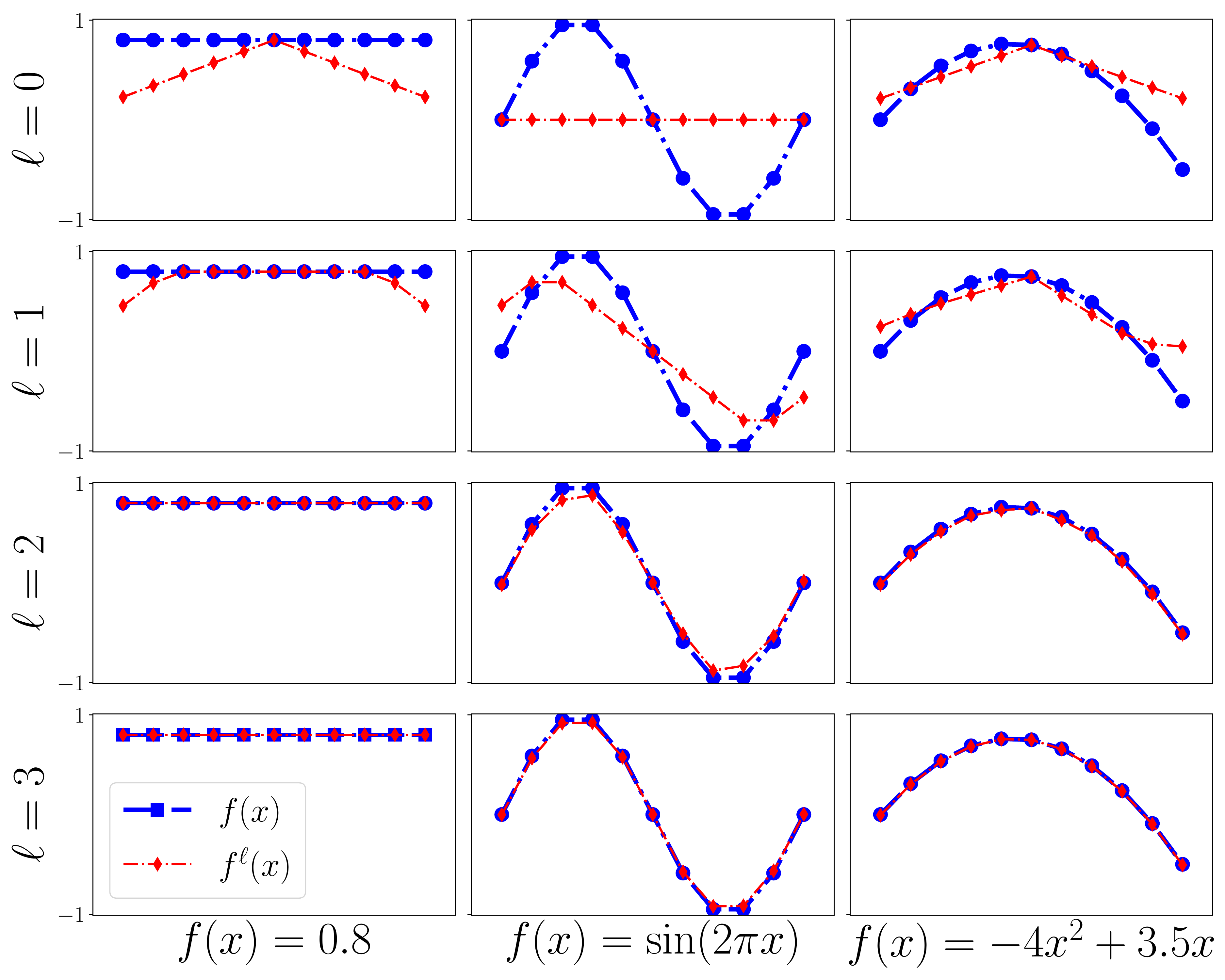}
    \caption{Interpolation on sparse grids with increasing resolution (i.e., $\ell$) for $1$-dimensional functions.
    }
\label{fig:hierarchical_sg_interpolation}
\end{figure}

We introduce $\mathbb{L}_{\ell, d} \definedas \{ (\bv{l}, \bv{i})  \mid  \bv{x}_{\bv{l}, \bv{i}} \in  \gridset_{\ell, d}  \}$ and $\bv{x}_{\bv{l}, \bv{i}} = \left[\bv{i}_1/2^{\bv{l}_1+1}, \cdots, \bv{i}_d/2^{\bv{l}_d+1} \right]$.
Then, for any pair of resolution and index vectors $(\bv{l}, \bv{i}) \in \mathbb{L}_{\ell, d}$, 
a tensorized hat function $\varphi_{\bv{l}, \bv{i}}(\bv{x})$\footnote{$\varphi_{\bv{l}, \bv{i}}(\bv{x}) \definedas \prod_{k=1}^{d} \varphi_{\bv{l}_k, \bv{i}_k}(\bv{x}_k)$ where  $\forall k, \varphi_{\bv{l}_k, \bv{i}_k}(\bv{x}_k) =  \varphi(\frac{\bv{x}_k - \bv{i}_k .2^{-\bv{l}_k}}{2^{-\bv{l}_k}})$ and $\varphi(x) \definedas \max
\{ 1 - |x|, 0  \} $. } is created such that it is centered at the location of the grid point corresponding to $(\bv{l}, \bv{i})$ and has support for a symmetric interval of length $2^{-\bv{l}_i}$ in the $i$ dimension. 
For any function $f: \R^{d} \mapsto R$, the sparse grid \textit{interpolant} rule $f^{\ell}: \R^{d} \mapsto R$ is given as:

\begin{align}
\label{eq:linear_interpolation_sparse_grid}
    f^{\mathcal{\ell}}({\bv x}) &=  \sum_{(\bv{l}, \bv{i}) \in \mathbb{L}_{\ell, d} }  \sum_{\bm{\delta} \in \Delta^d} (-2)^{- \norm{\bm{\delta}}_{0}} f(\bv{x}_{\bv{l}, \bv{i} + \bm{\delta}}) \varphi_{\bv{l}, \bv{i}} ({\bv x}) 
\end{align}
where $\Delta^d = \{-1, \,\,\, 0, \,\,\, 1\}^{d}$ is stencil evaluation of the function $f$ centered at grid-point  $\bv{x}_{\bv{l}, \bv{i}}$  \citep{pfluger2010spatiallyclassification}. Concretely,  $\bv{x}_{\bv{l}, \bv{i} + \bm{\delta}} \in \R^{d}$ has $k^{th}$ position equal to $ (\bv{i}_k+\bm{\delta}_k) 2^{-\bv{l}_k}$.
Figure \ref{fig:hierarchical_sg_interpolation} illustrates the above sparse grid \textit{interpolant} rule for simple 1-dimensional functions. 
Notice that it progressively gets more accurate as the resolution level $\ell$ increases.

Next, to interpolate the kernel function, 
we need $f^{\ell} (\bv{x})= \bv{w}_\bv{x}\theta$, where $\bv{w}_\bv{x} \in \R^{ 1 \times \cardinalitynorm{G^d_{\ell}}} $ and $  \theta \in \R^{\cardinalitynorm{G^d_{\ell}}}$ is the evaluation of $f$ on $\gridset_{\ell, d}$. 
As given such a formula, we can write $k(\bv{x}, \bv{x}') \approx \cov(f^{\mathcal{\ell}}({\bv x}), f^{\mathcal{\ell}}({\bv x}')) = \bv{w}_\bv{x}K_{\gridset_{\ell,d}} \bv{w}_\bv{x}^T $, where $K_{\gridset_{\ell,d}}$ is the true kernel matrix on sparse grid. 
Furthermore, by stacking interpolation weights $\bv{w}_\bv{x}$ into $W$ matrix for all data points similar to SKI, we can approximate the kernel matrix as $\tilde K_{X} = W  K_{\gridset_{\ell,d}}  W^T$.

\begin{mdframed}[backgroundcolor=light-gray]
\begin{claim}\label{clm:sparse_grid_gsgp} 
    $\forall \bv{x} \in \R^{d}, \ell \in \mathbb{N}_0, \exists \bv{w}_\bv{x} \,\,\, s.t. \,\,\, f^{\ell} (\bv{x})= \bv{w}_\bv{x}\theta$ 
    where $\theta \in \R^{\cardinalitynorm{\gridset^{\ell, d}}}$ is the evaluation of $f$ on $\gridset_{\ell, d}$
    , i.e, $\theta$  is made of $ \left\{ f(\bv{x}_{\bv{l}, \bv{i}}) \mid \bv{x}_{\bv{l}, \bv{i}} \in \gridset_{\ell, d}  \right\}$ and indexed to match the columns of $\bv{w}_\bv{x}\theta$.   
\end{claim}
\end{mdframed}

\begin{proof}

For brevity, we introduce, $Q_{\ell, d} \definedas \mathbb{L}_{\ell, d}  \setminus \bigcup_{0 \leq  l' \leq \ell-1} \mathbb{L}_{l', d}$, 
which is a partition of $\mathbb{L}_{\ell, d}$ based on the resolution of grid points, i.e., $\mathbb{L}_{\ell, d} = \bigcup_{0 \leq  l' \leq \ell} Q_{l', d} $.

\begin{align*}
    f^{\ell}({\bv x}) &=  \sum_{(\bv{l}, \bv{i}) \in \mathbb{L}_{\ell, d} }  \sum_{\bm{\delta} \in \Delta^d} (-2)^{- \norm{\bm{\delta}}_{0}} f(\bv{x}_{\bv{l}, \bv{i} + \bm{\delta}}) \varphi_{\bv{l}, \bv{i}} ({\bv x}) \\
       &= \sum_{0 \leq l' \leq \ell} \sum_{(\bv{l}, \bv{i}) \in Q_{l', d}} \left( \sum_{\bm{\delta} \in \{-1 \,\,\, 0 \,\,\, 1\}^d} 2^{- \norm{\bm{\delta}}_{0}} f(\bv{x}_{\bv{l}, \bv{i} + \bm{\delta}}) \right) \varphi_{\bv{l}, \bv{i}} ({\bv x}) \\
      &= \sum_{0 \leq l' \leq \ell} \sum_{(\bv{l}, \bv{i}) \in Q_{l', d}} \left( \sum_{\bm{\delta} \in \{-1 \,\,\, 0 \,\,\, 1\}^d} 2^{- \norm{\bm{\delta}}_{0}} f(\bv{x}_{\bv{l}, \bv{i} + \bm{\delta}}) \right) \varphi_{\bv{l}, \bv{i}} ({\bv x}) \\
      &= \sum_{0 \leq l' \leq \ell}  \sum_{(\bv{l}, \bv{i}^{'} - \bm{\delta}) \in Q_{l', d}}   \sum_{\bm{\delta} \in \{-1 \,\,\, 0 \,\,\, 1\}^d} 2 ^{- \norm{\bm{\delta}}_{0}} \varphi_{\bv{l}, \bv{i}^{'} - \bm{\delta}} ({\bv x}) f(\bv{x}_{\bv{l}, \bv{i}^{'}}) \,\,\, \text{ by substitution}  \,\,\, {\bv{i} = \bv{i}^{'} - \bm{\delta}} \\
      &= \sum_{0 \leq l' \leq \ell}  \left( \sum_{(\bv{l}, \bv{i}^{'} - \bm{\delta}) \in Q_{l', d}}   \sum_{\bm{\delta} \in \{-1 \,\,\, 0 \,\,\, 1\}^d} 2 ^{- \norm{\bm{\delta}}_{0}} \varphi_{\bv{l}, \bv{i}^{'} - \bm{\delta}} ({\bv x})  \right) f(\bv{x}_{\bv{l}, \bv{i}^{'}}) \\
      &= \sum_{0 \leq l' \leq \ell}  \left( \sum_{(\bv{l}, \bv{i}^{'}) \in Q_{l', d}}   \sum_{\bm{\delta} \in \{-1 \,\,\, 0 \,\,\, 1\}^d} 2 ^{- \norm{\bm{\delta}}_{0}} \varphi_{\bv{l}, \bv{i}^{'} + \bm{\delta}} ({\bv x})  \right) f(\bv{x}_{\bv{l}, \bv{i}^{'}}) \\ 
      &= \sum_{0 \leq l' \leq \ell}  \sum_{(\bv{l}, \bv{i}^{'}) \in Q_{l', d}}   \sum_{\bm{\delta} \in \{-1 \,\,\, 0 \,\,\, 1\}^d} 2 ^{- \norm{\bm{\delta}}_{0}} \varphi_{\bv{l}, \bv{i}^{'} + \bm{\delta}} ({\bv x})   f(\bv{x}_{\bv{l}, \bv{i}^{'}}) \\
      &= \sum_{0 \leq l' \leq \ell}   \sum_{(\bv{l}, \bv{i}) \in Q_{l', d}}   \sum_{\bm{\delta} \in \{-1 \,\,\, 0 \,\,\, 1\}^d} 2 ^{- \norm{\bm{\delta}}_{0}} \varphi_{\bv{l}, \bv{i} + \bm{\delta}} ({\bv x})   f(\bv{x}_{\bv{l}, \bv{i}}) \\
      &= \sum_{(\bv{l}, \bv{i}) \in \mathbb{L}_{\ell, d} }  \sum_{\bm{\delta} \in \{-1 \,\,\, 0 \,\,\, 1\}^d} 2 ^{- \norm{\bm{\delta}}_{0}} \varphi_{\bv{l}, \bv{i} + \bm{\delta}} ({\bv x})   f(\bv{x}_{\bv{l}, \bv{i}}) \\
\end{align*}

Though it may seem that not all $\varphi_{\bv{l}, \bv{i} + \bm{\delta}}$ are on 
the sparse grid as the components of $\bv{i}+ \bm{\delta}$ can be even. 
Fortunately, it is true as  $x_{l, i} = x_{l+1, 2i}$ by the construction of sparse grids, and we can apply the following transformation to uniquely project $(\bv{l}, \bv{i} + \bm{\delta})$ on $\gridset_{\ell, d}$ as follows:

\begin{equation*}
    \Delta(\bv{l}, \bv{i}, \bm{\delta}) = \left(\bv{l} - \#^{2}(\bv{i} + \bm{\delta}), \frac{\bv{i}}{2^{\#^{2}(\bv{i} + \bm{\delta})}}\right),
\end{equation*}

where, $\#^{2}$ computes the exponent of 2 in its prime factorization component-wise. Notice that output of $\Delta(\bv{l}, \bv{i}, \bm{\delta})$ is bound to be in $\gridset_{\ell, d}$  as the resultant-position index pair (i.e., $(\bv{l}, \bv{i})$) will have level index $l$ $\leq \ell$ and position index $i$ to be odd, for all components, respectively.


\begin{align*}
    f^{\ell}({\bv x})   &= \sum_{(\bv{l}, \bv{i}) \in \mathbb{L}_{\ell, d} } \underbrace{ \sum_{\bm{\delta} \in \{-1 \,\,\, 0 \,\,\, 1\}^d} (-2) ^{- \norm{\bm{\delta}}_{0}} \varphi_{\Delta(\bv{l}, \bv{i}, \bm{\delta})} ({\bv x}) }_{\definedas  \bv{w}_{\bv{l}, \bv{i}}(\bv{x}) }   f(\bv{x}_{\bv{l}, \bv{i}}) \numberthis{}
\end{align*}

By setting $\bv{w}_\bv{x}$ as $\bv{w}_{\bv{l}, \bv{i}}(\bv{x})$ from above equation for all $(\bv{l}, \bv{i}) \in \mathbb{L}_{\ell, d}$, 
we have $f^{\ell} (\bv{x})= \bv{w}_\bv{x}\theta$.

\end{proof}
 
\subsection{Combination Technique for Sparse Grid Interpolation}
\label{sssec:interpolation_via_combination_techique_sparse}

The combination technique \cite{combinationtechnique} provides yet another \textit{interpolant} rule for sparse grids. 
Formally, for any function $f: \R^{d} \mapsto R$, the sparse grid combination technique \textit{interpolant} rule $f^{\ell}_{c}: \R^{d} \mapsto R$ is:

\begin{equation}
\label{eq:combination_technique_interpolation}
    f^{\mathcal{\ell}}_{c}({\bv x}) =  \sum_{q=0}^{d-1}  (-1)^{q} {{d -1} \choose {q}}  \sum_{\{ \bv{l} \in \mathbb{N}_0^{d} \mid \cardinalitynorm{\bv{l}}_1 = \ell - q \}}  f_{\Omega_{\bv{l}}}(\bv{x}), 
\end{equation}

where, $f_{\Omega_{\bv{l}}}$ is an \textit{interpolant} rule on the rectilinear grid $\Omega_{\bv{l}}$ \citep{valentin2019b, combinationtechnique}. 
Consequently, extending simplicial and cubic interpolation from the rectilinear grid to sparse grids is trivial.
Notice that the interpolant $ f^{\mathcal{\ell}}_{c}$ uses a smaller set of grids, i.e., 
only  $\left\{\Omega_{\bv{l}} \mid \max\{\ell - d, 0\} <  \cardinalitynorm{\bv{l}}_1 \leq \ell \right\}$. 
Nevertheless, $ \forall \,\,\, \ell \,\,\, \& \,\,\, d$, the rectilinear grids used in $f^{\mathcal{\ell}}_{c}$ are always contained in $\gridset_{\ell, d}$.

Equation \ref{eq:combination_technique_interpolation} prototypes the construction of interpolation weights $W$ matrix.
 Concretely, interpolation weights are computed and stacked along columns for each dense grid $\Omega_{\bv{l}}$ in the sparse grid.
  After that columns are scaled by factor $(-1)^{q} {{d -1} \choose {q}}$ to satisfy $f^{\mathcal{\ell}}_{c}$.

 \section{Structured Kernel Interpolation on Sparse Grids -- Omitted details}
 \label{app:ski_on_sparse_grids}

 \subsection{Fast Multiplication with Sparse Grid Kernel Matrix}
 \label{app:fast_mvm_algorithm}

 \noindent \textbf{Indexing the kernel matrix $K_{\gridset_{\ell,d}}$}. 
 Recall \textsc{P3} from proposition \ref{proposition:sparse_grid_properties}, 
 i.e., the recursive construction of sparse grids, $\gridset_{\ell, d} = \bigcup {}_{i=0}^{\ell} \big(\Omega_i \otimes \gridset_{\ell-i, d-1}\big)$. We say $\gridset_{\ell, d}^i \definedas \Omega_i \otimes \gridset_{\ell-i, d-1}$.
 From \textsc{P3}, we know that $K_{\gridset_{\ell,d}}$ can be written as the block matrix such that 
 both rows and columns are indexed by all $\gridset_{\ell, d}^i$. 
 For all combinations of combinations $\gridset_{\ell, d}^i$, $the K_{\gridset_{\ell,d}}$ is a $(\ell + 1) \times (\ell +1)$ block matrix. 
 
 Similarly,  without the loss of generality, 
 any arbitrary vector  $\bv{v} \in \R^{\cardinalitynorm{\gridset_{\ell,d}}}$ 
 is indexed using $\gridset_{\ell, d}^i$ and the output vector after an MVM operation also 
 follows indexing by $\gridset_{\ell, d}^i$. Concretely, we let $\bv{u} = K_{\gridset_{\ell,d}} \bv{v}$, then $\forall \,\, 0 \le i \le \ell$, we can write $\bv{u}_i = \sum_{j=1}^{\ell} \tilde {\bv{v}}_{ij}$, where $\tilde {\bv{v}}_{ij} = K_{ \gridset_{\ell, d}^i,  \gridset_{\ell, d}^j} \bv{v}_j$, 
 rows of $\bv{v}$, rows of $\bv{u}$ and rows and columns of $K$, are indexed by the $\gridset_{\ell, d}^i$. 

\noindent \textbf{Structure and redundancy in the kernel sub-matrices}.
Note that $\gridset_{\ell, d}^i$ is the Cartesian product between rectilinear grid $\Omega_{i}$ and sparse grid $\gridset_{\ell-i, d-1}$ imposing Kronecker structure on the matrix $K_{ \gridset_{\ell, d}^i,  \gridset_{\ell, d}^j}$, given that $k$ is a product kernel. As a result, for each $\bv{u}_i$, 
we have:

\begin{equation} 
     \tilde {\bv{v}}_{ij} =  \vectorize  \left[ K_{\Omega_{i},\Omega_{j}} \matricize(\bv{v}_j) K_{\gridset_{\ell-i, d-1}, \gridset_{\ell-j, d-1}}^{T} \right]
    \label{eq:equation_of_vij}
\end{equation}
where $\vectorize$ and $\matricize$ are standard matrix reshaping 
operators used in multiplying vectors with a Kronecker product of two matrices. Observe that 
both $K_{\Omega_{i},\Omega_{j}}$ and $K_{\gridset_{\ell-i, d-1}, \gridset_{\ell-j, d-1}}$ are rectangular 
and have many common entries across different pairs of $i$ and $j$. For instance, 
$\forall j>i$, $ K_{\Omega_{i},\Omega_{j}} \subseteq K_{\gridset_{j,1}}$,
similarly,  we also have, 
$\forall j \le i, K_{\gridset_{\ell-i, d-1}, \gridset_{\ell-j, d-1}} \subseteq K_{\gridset_{\ell-i, d-1}}$ 
as $\gridset_{\ell-j, d-1} \subseteq \gridset_{\ell-i, d-1}$.

\noindent \textbf{Efficient ordering for Kronecker product and exploiting redundancy in kernel sub-matrices}.
Next, we observe the two different orders of computation for Equation \ref{eq:equation_of_vij}, 
i.e., first multiplying $K_{\Omega_{i},\Omega_{j}}$ with versus multiplying with $K_{\gridset_{\ell-i, d-1}, \gridset_{\ell-j, d-1}}^{T}$.
To exploit this choice and leverage the redundancy mentioned above, we divide the computation as below\footnote{This is in part inspired by \citet{fastmvmgalerkinmethod} as our algorithm also orders computation by first dimension of sparse grid (i.e., $\gridset_{\ell, d}^i)$.}:

\begin{align} 
    \label{eq:frgamentation}
    \bv{u}_i = \bv{a}_i + \bv{b}_i, \,\,\,\, \text{where}, \,\,\,\, \bv{a}_i = \sum_{j>i}^{\ell} \tilde {\bv{v}}_{ij}, \,\,\,\,  \text{and} \,\,\,\,  \bv{b}_i = \sum_{j=0}^{i} \tilde {\bv{v}}_{ij}.     
\end{align}

In Algorithm \ref{alg:sgkma_algorithm}, $A_i$ and $B_i$ 
are such that $\vectorize(A_i) = \bv{a}_i$ 
and $\vectorize(A_i) = \bv{b}_i$.

\begin{mdframed}[backgroundcolor=light-gray]
\begin{claim}\label{clm:computaiton_a_i} With ${\overline{A}}_i \definedas K_{\gridset_{i,1}} \mathcal{S}_{\gridset_{i,1}, \Omega_{i}} \matricize(\bv{v})$, $\forall 0 \leq i \leq \ell$,  $\bv{a}_i$ can be given as follows: 
\begin{align*}
\bv{a}_i = \vectorize \left[ \left(\sum_{j>i} 
S_{\Omega_{i},\gridset_{j,1}} {\overline{A}}_j \mathcal{S}_{\gridset_{\ell-j,d-1}, \gridset_{\ell-i,d-1}} \right) K_{\gridset_{\ell-i,d-1}} \right]
\end{align*}
\end{claim}
\end{mdframed}

\begin{proof}
\begin{align*}
    \bv{a}_i &= \sum_{j>i}^{\ell} \tilde {\bv{v}}_{ij} \\ 
    &=\sum_{j>i}^{\ell}   \vectorize  \left[ K_{\Omega_{i},\Omega_{j}} \matricize(\bv{v}_j) K_{\gridset_{\ell-i, d-1}, 
    \gridset_{\ell-j, d-1}}^{T} \right]  \text{(from the Equation  \ref*{eq:equation_of_vij})}\\
    &=\sum_{j>i}^{\ell} \vectorize  \left[ K_{\Omega_{i},\Omega_{j}} \matricize(\bv{v}_j)  \mathcal{S}_{\gridset_{\ell-j, d-1}, \gridset_{\ell-i, d-1}}  K_{\gridset_{\ell-i,d-1}}  \right]  \text{(by expanding the kernel matrix)}\\
    &=\vectorize  \left[ \sum_{j>i}^{\ell}  \left( K_{\Omega_{i},\Omega_{j}} \matricize(\bv{v}_j)  \mathcal{S}_{\gridset_{\ell-j, d-1}, \gridset_{\ell-i, d-1}} \right)  K_{\gridset_{\ell-i,d-1}}  \right] \text{(using linearity of operations)}\\
    &=\vectorize  \left[ \sum_{j>i}^{\ell}  \left( K_{\Omega_{i},\gridset_{j,1}} \mathcal{S}_{\gridset_{j,1}, \Omega_{j}} \matricize(\bv{v})  \mathcal{S}_{\gridset_{\ell-j, d-1}, \gridset_{\ell-i, d-1}} \right)  K_{\gridset_{\ell-i,d-1}}  \right] \text{(by expanding kernel matrix)} \\
    &=\vectorize  \left[ \sum_{j>i}^{\ell}  \left(\mathcal{S}_{\Omega_{i}, \gridset_{j,1}}  K_{\gridset_{j,1}} \mathcal{S}_{\gridset_{j,1}, \Omega_{j}} \matricize(\bv{v})  \mathcal{S}_{\gridset_{\ell-j, d-1}, \gridset_{\ell-i, d-1}} \right)  K_{\gridset_{\ell-i,d-1}}  \right]  \\
    &=\vectorize  \left[ \sum_{j>i}^{\ell}  \left(\mathcal{S}_{\Omega_{i}, \gridset_{j,1}}  \overline{A}_j \mathcal{S}_{\gridset_{\ell-j, d-1}, \gridset_{\ell-i, d-1}} \right)  K_{\gridset_{\ell-i,d-1}}  \right]  \text{(enables pre-computation using $\overline{A}_j$)}
\end{align*}
\end{proof}


Intuitively, the claim \ref{clm:computaiton_a_i} demonstrates the rationale behind
the operator $\mathcal{S}$, 
because (1) it reduces the computation from $\frac{\ell^2}{2}$ MVM with $K_{\Omega_{i},\Omega_{j}}$ to only $\ell$ MVM with $ K_{\gridset_{i,1}}$ 
Toeplitz matrices via pre-computing ${\overline{A}}_i$, and (2) it requires only $\ell$ MVM with $K_{\gridset_{\ell-i,d-1}}$ instead of $\frac{\ell^2}{2}$ MVMs with $K_{\gridset_{\ell-i, d-1}, \gridset_{\ell-j, d-1}}$ via exploiting 
linearity of operations involved.  Following analogous steps, 
$\bv{b}_i$ can be derived $\bv{b}_i = \vectorize(B_i)$.

\begin{mdframed}[backgroundcolor=light-gray]
\begin{claim}\label{clm:computaiton_b_i} With ${\overline{B}}_i \definedas V_i K_{\gridset_{\ell-i, d-1}} $, $\forall 0 \leq i \leq \ell$,  $\bv{b}_i$ can be given as follows: 
\begin{align*} 
    \bv{b}_i &=  \vectorize \left[ \mathcal{S}_{\Omega_i, \gridset_{i, 1}} K_{\gridset_{i,1}}  
    \left( \sum\limits_{j \leq i} \mathcal{S}_{\gridset_{i,1}, \Omega_j}  {\overline{B}}_{j} \mathcal{S}_{\gridset_{\ell- j, d-1}, \gridset_{\ell - i, d-1}} \right) \right]
\end{align*}
\end{claim}
\end{mdframed}        

\begin{proof}
\begin{align*}
    \bv{b}_i &= \sum_{j \leq i} \tilde {\bv{v}}_{ij} \\   
    &= \sum_{j \leq i} \ \vectorize  \left[ K_{\Omega_{i},\Omega_{j}} \matricize(\bv{v}_j) K_{\gridset_{\ell-i, d-1}, \gridset_{\ell-j, d-1}}^{T} \right]  \text{(from Equation  \ref*{eq:equation_of_vij})}\\   
    &= \sum_{j \leq i} \ \vectorize  \left[ K_{\Omega_{i},\Omega_{j}}  V_j K_{\gridset_{\ell-j, d-1}}  \mathcal{S}_{\gridset_{\ell-j, d-1}, \gridset_{\ell-i, d-1}} \right] \\
    &= \sum_{j \leq i} \ \vectorize  \left[ \mathcal{S}_{\Omega_i, \gridset_{i, 1}} K_{\gridset_{i,1}, \Omega_{j}}  V_j K_{\gridset_{\ell-j, d-1}}  \mathcal{S}_{\gridset_{\ell-j, d-1}, \gridset_{\ell-i, d-1}} \right] \\
    &= \sum_{j \leq i} \ \vectorize  \left[ \mathcal{S}_{\Omega_i, \gridset_{i, 1}} K_{\gridset_{i,1}}  \mathcal{S}_{\gridset_{i, 1}, \Omega_{j}} V_j K_{\gridset_{\ell-j, d-1}}  \mathcal{S}_{\gridset_{\ell-j, d-1}, \gridset_{\ell-i, d-1}} \right] \\
    &=  \vectorize \left[ \mathcal{S}_{\Omega_i, \gridset_{i, 1}} K_{\gridset_{i,1}}  \left( \sum\limits_{j \leq i} 
    \mathcal{S}_{\gridset_{i,1}, \Omega_j}  {\overline{B}}_{j} \mathcal{S}_{\gridset_{\ell- j, d-1}, \gridset_{\ell - i, d-1}} \right) \right]
\end{align*}
\end{proof}




\begin{mdframed}[backgroundcolor=light-gray]
\begin{reptheorem}{thm:sparse_grid_mvm_algorithm} 
\change{
    Let $K_{\gridset_{\ell,d}}$ be the kernel matrix for 
    a $d$-dimensional sparse grid with resolution $\ell$ for 
    a stationary product kernel. For any $\bv v \in \mathbb{R}^{\cardinalitynorm{\gridset_{\ell,d}}}$, 
  Algorithm~\ref{alg:sgkma_algorithm} computes $K_{\gridset_{\ell,d}}\bv v$ 
  in $\O(\ell^{d} 2^\ell)$ time.
    }
\end{reptheorem}
\end{mdframed}

\begin{proof}

\noindent \textbf{The correctness of the Algorithm \ref*{alg:sgkma_algorithm}}. 
Equation \ref{eq:frgamentation}, 
claim \ref{clm:computaiton_a_i} and claim \ref{clm:computaiton_b_i} establish 
the correctness of the output of Algorithm \ref{alg:sgkma_algorithm}, i.e., it computes $K_{\gridset_{\ell,d}}\bv v$.

\noindent \textbf{On the complexity of Algorithm \ref*{alg:sgkma_algorithm}}. (We prove it by induction on $d$.) \hfill \break

\noindent \textbf{Base case:} For any $\ell$ and $d=1$, the algorithm utilizes Toeplitz multiplication 
which require only  $\cardinalitynorm{\gridset_{\ell, 1}} \log \cardinalitynorm{\gridset_{\ell, 1}}$,
as $\cardinalitynorm{\gridset_{\ell, 1}} = 2^{\ell + 1}$, total required computation is $\O(\ell 2^\ell)$. 

\noindent \textbf{Inductive step:} We assume that the complexity holds, i.e., $\O(\ell^{d} 2^\ell)$ for $d-1$, 
then it's sufficient to show that Algorithm \ref{alg:sgkma_algorithm} needs only $\O(\ell^{d} 2^\ell)$ for $d$, in order to complete the proof.
Below, we establish the same separately for both pre-computation steps (i.e., Line 6 to 9) and the main loop (i.e., Line 11 to 15) of Algorithm \ref{alg:sgkma_algorithm}. Before that, we state an important fact for 
the analysis of the remaining steps:
\begin{align}\label{eq:useful_relation}
\sum_{i=0}^{\ell} \cardinalitynorm{\gridset_{i, 1}}  \times  \cardinalitynorm{\gridset_{\ell - i, d-1}} = \sum_{i=0}^{\ell} 2 \times \cardinalitynorm{\Omega_{i}}  \times  \cardinalitynorm{\gridset_{\ell - i, d-1}} \underbrace{=}_{\textsc{P3}} 2 \cardinalitynorm{\gridset_{\ell, d}} =  O(\ell^{d-1} 2^\ell)
\end{align}

\textbf{Analysis of the pre-computation steps}.
\begin{itemize}
   \item For reshaping $\bv{v}$ into $V_i$'s, we need  $\sum_{i=0}^{\ell} \cardinalitynorm{\Omega_{i}} \times \cardinalitynorm{\gridset_{\ell-i, d-1}} = \cardinalitynorm{\gridset_{\ell, d}} = O(\ell^{d-1} 2^\ell)$.
   \item For the rearrangement $V_i$ into $\mathcal{S}_{\gridset_{i,1}, \Omega_i} V_i$, we need $\sum_{i=0}^{\ell} \cardinalitynorm{\gridset_{i, 1}} \times \cardinalitynorm{\gridset_{\ell-i, d-1}} $ as operator $\mathcal{S}$ maps $V_i$ directly into the result.
   Therefore, we need $O(\ell^{d-1} 2^\ell)$ using Equation \ref{eq:useful_relation}.
   \item For the ${\overline{A}}_i$ step:
   \begin{itemize}
       \item $\forall i$, $\cardinalitynorm{\gridset_{\ell - i, d-1}}$ vectors are multiplied with Toeplitz matrix of size $\cardinalitynorm{\gridset_{i, 1}} \times \cardinalitynorm{\gridset_{i, 1}}$,
       \item so total computation for line $7$ is, $\sum_{i=0}^{\ell} \cardinalitynorm{\gridset_{i, 1}} \log \cardinalitynorm{\gridset_{i, 1}}  \times  \cardinalitynorm{\gridset_{\ell - i, d-1}}  
 = \sum_{i=0}^{\ell} i \times \cardinalitynorm{\gridset_{i, 1}}  \times  \cardinalitynorm{\gridset_{\ell - i, d-1}} \leq \sum_{i=0}^{\ell} \ell \times \cardinalitynorm{\gridset_{i, 1}}  \times  \cardinalitynorm{\gridset_{\ell - i, d-1}}  \underbrace{\leq}_{\text{Eq. \ref{eq:useful_relation}}}  
 \ell \times  \O(\ell^{d-1} 2^\ell) =  \O(\ell^{d} 2^\ell)$.
   \end{itemize}
   \item For the ${\overline{B}}_i$ step:
   \begin{itemize}
       \item $\forall i$, $\cardinalitynorm{\Omega_i}$ vectors need to be multiplied with $K_{\gridset_{\ell - i, d-1}}$, 
       \item using induction, total computation for line $8$ is, $\sum_{i=0}^{\ell} 2^i \times (\ell - i)^{d} 2^{\ell - i} = 2^{\ell} \sum_{i=0}^{\ell}(\ell - i)^{d} \leq 2^{\ell} \sum_{i=0}^{\ell}\ell ^{d-1} =  \O(\ell^{d} 2^\ell)$.
   \end{itemize}
\end{itemize}

\textbf{Analysis of the main loop}.
\begin{itemize}
    \item For the $A_i$ step, 
    \begin{itemize}
        \item  $\ell-i$ rearrangements and summations (i.e., $\sum\limits_{j > i} \mathcal{S}_{\Omega_i, \gridset_{j,1}} {\overline{A}}_{j} \mathcal{S}_{\gridset_{\ell-j,d-1},\gridset_{\ell-i, d-1}}$) are performed 
        simultaneously (i.e., by appropriately summing ${\overline{A}}_{j}$ to the final result),
        \item therefore, the total computation for rearrangements and summation is, $\sum_{i=0}^{\ell} (\ell - i) \times \cardinalitynorm{\Omega_i} \times \cardinalitynorm{\gridset_{\ell-i, d-1}} 
        \leq \ell \sum_{i=0}^{\ell}  \cardinalitynorm{\Omega_i} \times \cardinalitynorm{\gridset_{\ell-i, d-1}} \underbrace{=}_{\textsc{P3}} \ell \times  \O(\ell^{d-1} 2^\ell)  = \O(\ell^{d} 2^\ell)$;
        \item $\forall i$, $\cardinalitynorm{\Omega_i}$ vectors need to be multiplied with $K_{\gridset_{\ell - i, d-1}}$, which is same computation as used in the $\bv{\overline{b}}_i$ step, therefore, it is  $\O(\ell^{d+1} 2^\ell)$.
    \end{itemize}
    \item For the $B_i$ step,
    \begin{itemize}
        \item similar to $A_i$, $i$ rearrangements and summation (required for 
        $\sum\limits_{j \leq i} \mathcal{S}_{\gridset_{i,1}, \Omega_j}  {\overline{B}}_{j} \mathcal{S}_{\gridset_{\ell- j, d-1}, \gridset_{\ell - i, d-1}}$), are performed simultaneously,
        \item therefore, total computation for rearrangements and summation is, $\sum_{i=0}^{\ell} i \times \cardinalitynorm{\gridset_{i, 1}} \times \cardinalitynorm{\gridset_{\ell-i, d-1}} \underbrace{\leq}_{\text{Eq. \ref{eq:useful_relation}}}  
       \ell \times \O(\ell^{d-1} 2^\ell) = \O(\ell^{d} 2^\ell)$.
        \item the total MVM computation with $K_{\gridset_{\ell, 1}}$ is same as for the $\overline{A}_i$ step, therefore its $\O(\ell^{d} 2^\ell)$, as shown earlier. 
    \end{itemize} 
    \item Finally, for the last re-arrangement in line $14$,  all updates are accumulated on $\bv{u}_i$. All $\bv{u}_i$ jointly are as large as $\cardinalitynorm{\gridset_{\ell, d}} $, therefore $\O(\ell^{d-1} 2^\ell)$ computation is sufficient.
\end{itemize}
\end{proof}

\paragraph*{Batching-efficient Reformulation of Algorithm \ref*{alg:sgkma_algorithm}.} \hfill \break
\label{app:iterative_implementation}

In short, the main ideas behind iterative implementation can be summarized below:
\begin{itemize}
    \item The recursions in Lines 8 and 12 can be batched together. 
    \item Similarly, the recursion spawns many recursive multiplications with kernel matrices of the form $\gridset_{\ell', d'}$ for $0 \leq \ell' < \ell$ and $1 \leq d' < d$, 
\end{itemize}

To achieve the above, we make the following modifications:
\begin{itemize}
    \item Re-organize computation of the Algorithm \ref{alg:sgkma_algorithm} and first loop over to compute $\overline{A}_i$ and $A_i$, followed by second loop over $\overline{B}_i$ and $B_i$.
    \item Notice since the computation of $\bv{u}_i$ depends on $B_i$, it implies that kernel-MVM with remaining dimensions need to be computed. Therefore, we run the second loop 
    over $\overline{B}_i$ and $B_i$ in the reverse order of dimensions compared to Algorithm \ref{alg:sgkma_algorithm}.
    \item At all computation steps, vectors are appropriately batched before multiplying with kernel matrices to improve efficiency. 
\end{itemize}

\subsection{Simplicial Interpolation on Rectilinear Grids -- Omitted details}
\label{app:simplicial_interpolation_onrectlinear_grids} 

For a detailed exposition of simplicial interpolation with rectilinear grids, we refer readers to \citet{halton1991simplicial}. 
The main idea is that each hypercube is partitioned into simplices, so the grid points themselves are still on the rectilinear grid 
(i.e., the corners of the hypercubes). For each grid point, the associated basis function 
takes value $1$ at the grid point and is non-zero only for the simplices 
adjacent to that point and takes value $0$ at the corner of those simplices. 
Therefore, it is linear on each simplex.

Concretely, for any $\bv{x} \in \R^{d}$, following steps are used to find basis 
function values and find grid points rectilinear grids that form the simplex containing $\bv{x}$.

\begin{enumerate}
    \item Compute local coordinates $\bv{r}^{\bv{x}} = \left[\lambda(\bv{x}_1, \bv{s}_1), \cdots \lambda(\bv{x}_d, \bv{s}_d) \right]$, where $\forall i, \lambda(\bv{x}_i, \bv{s}_i) = \bv{x}_i - \lfloor \bv{x}_i/\bv{s}_i \rfloor \bv{s}_i $ and 
    $\bv{s} \in \R^{d}$ is the spacing of the rectilinear grid, i.e., the distance between adjacent grid points along all dimensions.
    \item Sort local coordinates. We put $\bv{r}_{\bv{x}}$ in non-decreasing order, i.e., $\left\{\bv{o}_1, \bv{o}_2, \cdots, \bv{o}_d\right\} = \left\{1, 2, \cdots, d\right\} $ such that $1 \geq \bv{r}^{\bv{x}}_{\bv{o}_1}\ \geq \bv{r}^{\bv{x}}_{\bv{o}_2}\ \geq \cdots \geq \bv{r}^{\bv{x}}_{\bv{o}_d} \geq 0$ holds.
    \item Compute interpolating basis values $\bv{v}_{\bv{x}} \in \R^{d+1}$ as $\left[1- \bv{r}^{\bv{x}}_{\bv{o}_1}, \bv{r}^{\bv{x}}_{\bv{o}_1} - \bv{r}^{\bv{x}}_{\bv{o}_2}, \bv{r}^{\bv{x}}_{\bv{o}_2} - \bv{r}^{\bv{x}}_{\bv{o}_3}, \cdots, \bv{r}^{\bv{x}}_{\bv{o}_{d-1}} - \bv{r}^{\bv{x}}_{\bv{o}_d}, \bv{r}^{\bv{x}}_{\bv{o}_d} \right]$.  
    \item Obtain neighbors by sorting the coordinates (columns) of
     the reference simplex (described below) to follow the same sorting order as the local coordinates. I.e., we sort the reference coordinates by
      the inverse sorting of the local coordinates.
\end{enumerate}

Recall from the main text that there are several ways to partition the hypercube, i.e., several choices to build reference simplex.  
We build reference simplex $S \in \R^{d+1, d}$ by stacking $d+1$ row vectors, in particular, $\bv{1}^{p} \in \R^{1 \times d}$ vectors for $p \in \left[0, d\right]$ are stacked, 
where $\bv{1}^{p} \in \{0, 1\}^{d}$ has $d-p$ zeros followed by ones for the left-over entries.


\section{Experiments -- Omitted details and more results}
\label{app:experiments}

\subsection{Hyperparameters, optimization, and data processing details}
\label{app:hyperparameters_details}

We run our experiments on Quadro RTX 8000 with $48$ GB of memory.
For all experiments, we have used RBF kernel with separate length-scale for each dimension. 
For the optimization marginal log-likelihood, we use Adam optimizer with a learning rate $0.1$ for $100$ number of epochs. 
The optimization is stopped if no improvement is observed in the log-likelihood for $5$ consecutive epochs.

The CG train and test tolerance are set to $1.0$ and $0.01$, which do not worsen performance in practice. 
Both CG pre-conditioning rank and maximum are 100. Our data is split in the ratio of $4:2:3$ to form the train, validation, and test splits. 
All UCI datasets are standardized using the training data to have zero mean and unit variance.
For sparse-grid, we explore $ \ell \in [2, 3, 4, 5]$ for Table \ref{table:rmse_on_uci_datasets}. 
For dense-grid with simplicial interpolation, we explored grid points per dimension until we ran out of memory.




\subsection{Another interpolation rule to apply sparse grids to large scale dataset}

Recall that the relatively higher number of rectilinear grids used in a sparse grid slows them down on large-scale datasets. 
Analogous to the combination rule, we devise a new interpolation rule that only considers rectilinear grids in $\{ \Omega_{\bv{l}} \mid \norm{\bv{l}}_1 = \ell \mid  
\left( \ell \in \bv{l} \,\,\,  \textsc{or} \,\,\, \ell-1 \in \bv{l} \right) \}$, i.e., $d^2/2 + d$ grids.
Similar to the combination interpolation technique, all grid interpolation weights are scaled by one by the total number of grids considered.

We focus on two large datasets with relatively higher dimensions: Houseelectric and Airline. 
House electric has $\approx 2.05$ million data points with dimensionality $d=11$. 
Similarly, the Airline dataset has $\approx 5.92$ million data points with dimensionality $d=8$.
For Houseelectric, Sparse-grid performs comparably to Simplex-GP while being $3.95$x faster. 
SKIP and SGPR are out of memory for the airline dataset, while Simplex-GP is slower by more than $4$ orders of magnitude.
These results show that sparse grids with simplicial interpolation can be effective and efficient for large-scale datasets.

\begin{table}[]
\caption{Test root-mean-square error (RMSE) and inference time on two large datasets
with dimensions $d \geq 8$ and $n \geq 1M$. See text for more details on datasets.
All numbers are averaged over three trials. $\textsc{OOM}$ is out of memory. 
$\star$ number is taken from \citet{SimplicesGP}.}
\centering
\begin{tabular}{|l|c|c|c|c|}
\hline
Methods  & \multicolumn{2}{c|}{Houseelectric} &\multicolumn{2}{c|}{Airline}\\ \hline
   & RMSE & Time (in secs)  & RMSE & Time (in secs)  \\ \hline
SGPR   & $0.067^{\star}$ & -  & OOM  & - \\ \hline
SKIP  & OOM  & - & OOM  & - \\ \hline
Simplex-GP   &  $\bf{0.078}$ & $0.186$   & $0.922$ & $142.891$ \\ \hline        
Dense-grid   & $0.170$ & $0.263$ & $0.892$  & $0.413$ \\ \hline
Sparse-grid  &  $\bf{0.088}$ &  $\bf{0.047}$ & $\bf{0.832}$ & $\bf{0.003}$ \\ \hline
\end{tabular}
\label{table:rmse_on_uci_datasets_large}
\end{table}

\subsection{Sparse grid interpolation and GP inference for more synthetic functions.}
\label{app:synthetic_interpolation_gp_inference}

\begin{figure}[H]
    \begin{subfigure}{0.45 \textwidth}
    \centering \includegraphics[width=\textwidth] 
    {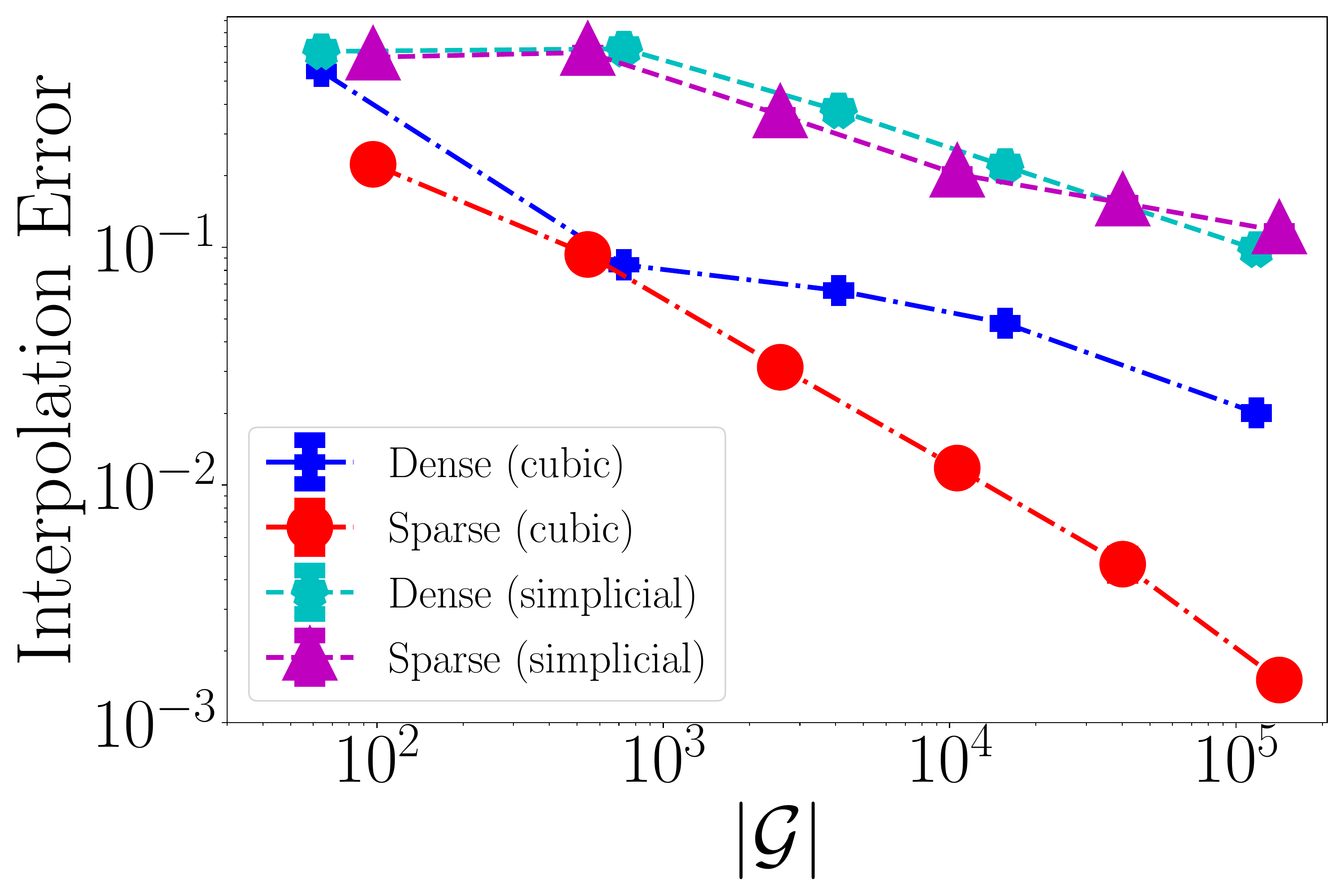}
    \end{subfigure}
    \hfill
    \begin{subfigure}{0.45 \textwidth}
    \centering \includegraphics[width=\textwidth] 
    {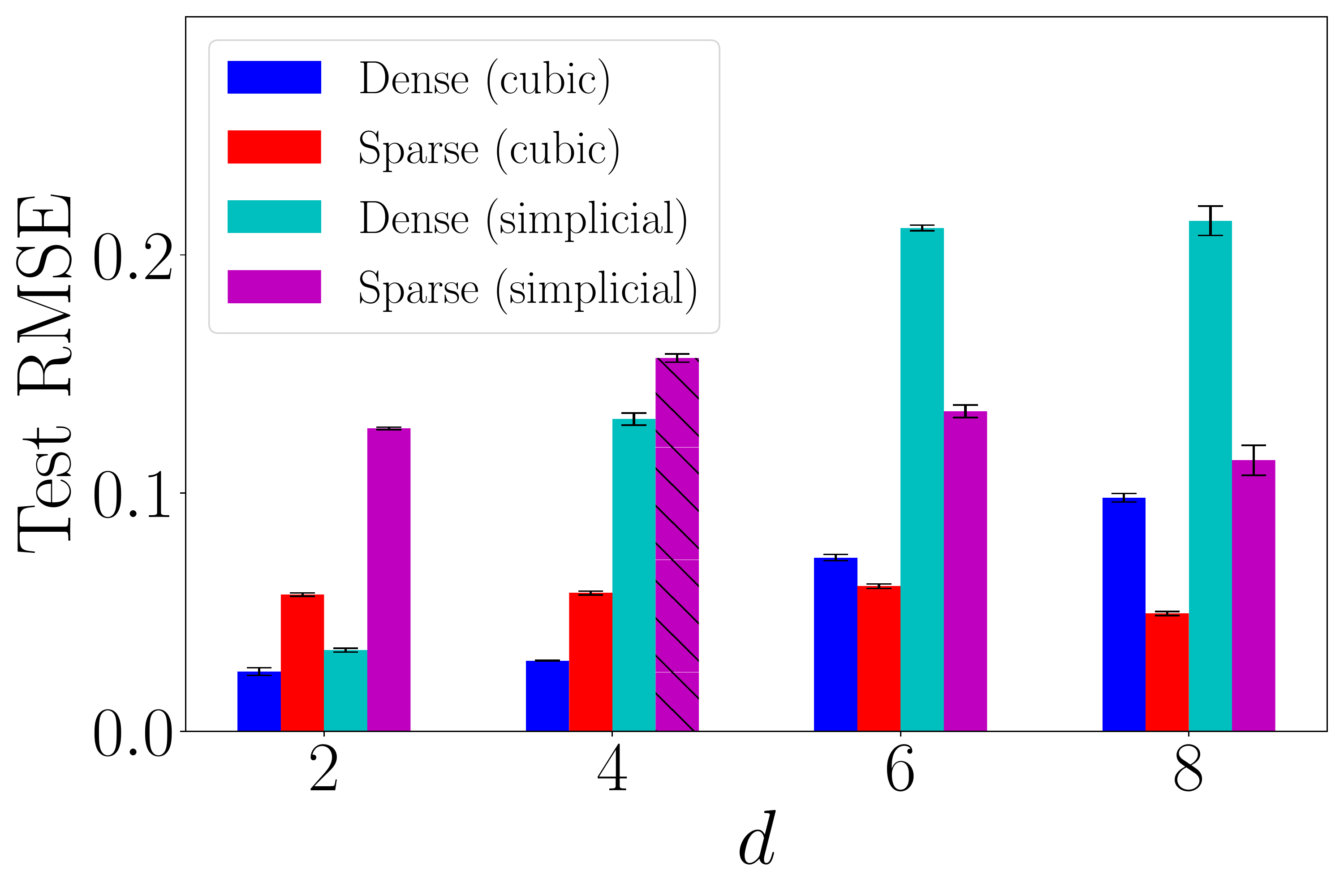} 
    \end{subfigure}
    \begin{subfigure}{0.45 \textwidth}
      \centering \includegraphics[width=\textwidth] 
      {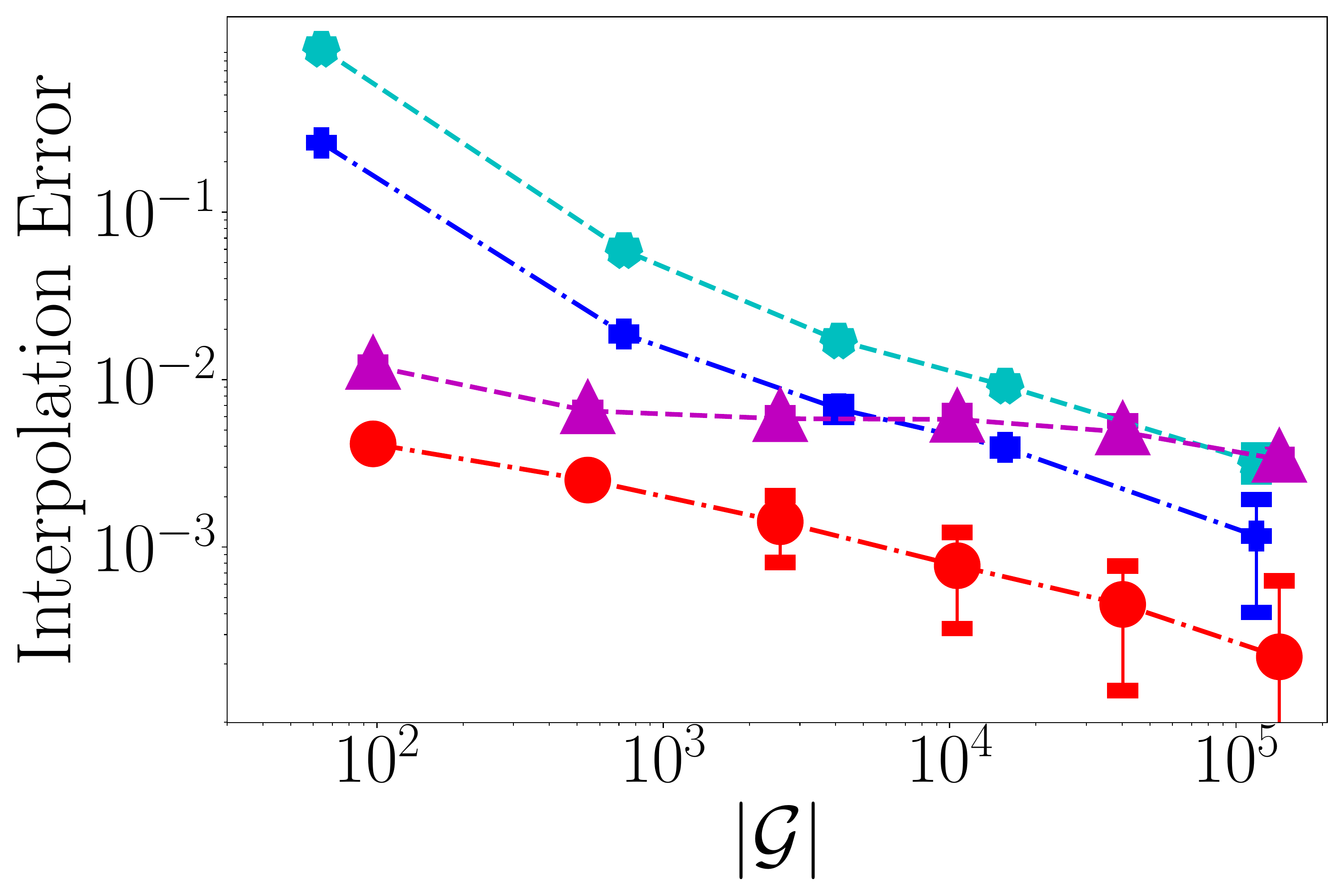}
      \end{subfigure}
      \hfill
      \begin{subfigure}{0.45 \textwidth}
      \centering \includegraphics[width=\textwidth] 
      {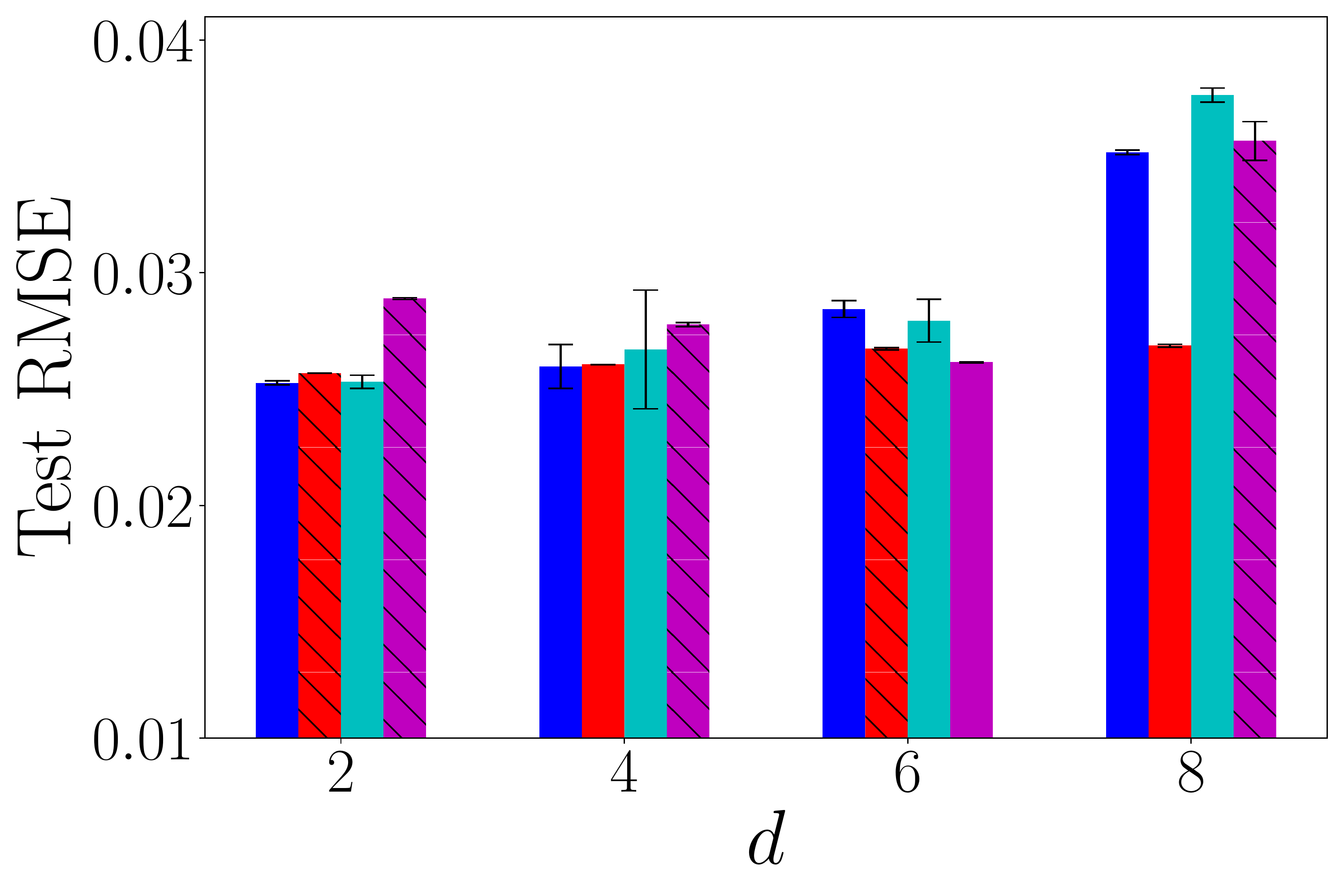} 
      \end{subfigure}
    \caption{Additional results for two more synthetic functions, namely, \textit{anisotropic and shifted} cosine (top panels) and \textit{corner-peak} (bottom panels). 
      Both left figures show that the interpolation accuracy of sparse grids is comparable or superior for both interpolation schemes (i.e., cubic and simplicial). 
    Furthermore, both right figures show that the advantage of sparse grids becomes more prominent as dimension increases. 
    This effect is relatively less prevalent in the bottom panel as the \textit{corner-peak} function attains smaller values with an increase in dimension.
    See text for precise function definitions. 
    }
 \label{fig:function_interpolation_gp_task_11}
 \end{figure}

Similar to section \ref{sec:experiments}, we consider two more functions that are not isotropic (unlike $\cos(\norm{\bv{x}}_1)$): 
a) \textit{anisotropic and shifted} cosine function $f_d^{as}(\bv{x}) \definedas \cos(2\pi w + \sum_{i=1}^{d}  \bv{x}_i \bv{c}_i)$, 
b) \textit{corner-peak} function $f_d^{cp}(\bv{x}) \definedas (1 + \sum_{i=1}^{d}  \bv{x}_i \bv{c}_i)^{-d-1}$. 
Both $w$ and $\bv{c}$ are selected randomly, and the remaining settings (i.e., noise, train, and evaluation procedures) are the same as in section \ref{sec:experiments}. 

\end{appendices}

\end{document}